\documentclass[5p,times]{elsarticle}

\usepackage{algorithm, setspace}
\usepackage{algorithmic}

\usepackage{multicol}
\usepackage{multirow}
\usepackage[table]{xcolor}
\usepackage{tcolorbox}
\usepackage{multirow}
\usepackage{tikz}
\usepackage{wrapfig}
\usepackage{graphicx}
\usepackage{url}
\usepackage{xfrac}

\AtBeginDocument{%
  }

\newcommand{\cir}[1]{\tikz[baseline]{%
    \node[anchor=base, fill=darkgray, draw, circle, inner sep=0, minimum width=1.1em]{#1};}}

\newcommand{\bailey}[1]{\textcolor{black}{#1}}



\journal{Future Generation Computer Systems}

\newcommand{\system}{\texttt{\textsc{Mosaic}}}

\begin{document}
\sloppy

\begin{frontmatter}


\title{Mosaic: Composite Projection Pruning for Resource-efficient LLMs}

\author[sta]{Bailey J. Eccles\corref{cor1}}
\ead{bje1@st-andrews.ac.uk}
\author[rmi]{Leon Wong}
\ead{leon.wong@rakuten.com}
\author[sta]{Blesson Varghese}
\ead{blesson@st-andrews.ac.uk}

\cortext[cor1]{Corresponding author}

\affiliation[sta]{organization={School of Computer Science, University of St Andrews},
            addressline={Jack Cole Building}, 
            city={St Andrews},
            postcode={KY16 9SX}, 
            state={Fife},
            country={Scotland, United Kingdom}}

\affiliation[rmi]{organization={Autonomous Networking Research \& Innovation Department, Rakuten Mobile, Inc.},
            addressline={Rakuten Crimson House, 1-14-1 Tamagawa, Setagaya-ku}, 
            city={Tokyo},
            postcode={158-0094}, 
            state={Tokyo},
            country={Japan}}

\begin{abstract}
Extensive compute and memory requirements limit the deployment of large language models (LLMs) on any hardware. 
Compression methods, such as pruning, can reduce model size, which in turn reduces resource requirements. 
State-of-the-art pruning is based on coarse-grained methods. 
They are time-consuming and inherently remove critical model parameters, adversely impacting the quality of the pruned model. 
This paper introduces projection pruning, a novel fine-grained method for pruning LLMs. 
In addition, LLM projection pruning is enhanced by a new approach we refer to as composite projection pruning - the synergistic combination of unstructured pruning that retains accuracy and structured pruning that reduces model size.
We develop \system{}, a novel system to create and deploy pruned LLMs using composite projection pruning. 
\system{} is evaluated using a range of performance and quality metrics on multiple hardware platforms, LLMs, and datasets.
\system{} is 7.19$\times$ faster in producing models than existing approaches.
\system{} models achieve up to 84.2\% lower perplexity and 31.4\% higher accuracy than models obtained from coarse-grained pruning. 
Up to 67\% faster inference and 68\% lower GPU memory use is noted for \system{} models. \system{} is available for public use from \url{https://github.com/blessonvar/Mosaic}
\end{abstract}


\end{frontmatter}

\section{Introduction}
Large language models (LLMs), such as GPT-4~\cite{gpt4}, have found applications in chatbots~\cite{bert} and generating content~\cite{gpt3}. LLMs consist of several billion~\cite{llama} to hundreds of billions~\cite{gpt3} of parameters. Consequently, training and deploying these LLMs on even state-of-the-art hardware is challenging due to their high memory and compute demands. For example, GPT-3, a 175 billion parameter LLM, is over 350 GB in size and requires $3.14 \times 10^{23}$ flops to train~\cite{gpt3} and five 80 GB Nvidia A100 GPUs to deploy. 
LLMs, such as ChatGPT~\cite{gpt4} and Gemini~\cite{gemini}, are hosted on clusters with tens of thousands of GPUs. Hence, LLMs are trained and served from cloud data centers where their resource demands can be met. 

In this context, we note the following two avenues within LLM research:

(1) \textbf{Lowering resource demands for LLM inference}. 
Serving LLMs on relatively resource-limited edge/mobile environments is challenging~\cite{llm-edge, phi}, given the substantial resource requirements of LLMs that exceed the available hardware in these environments. 
Running LLMs suited for such environments will reduce the need to send queries outside a user-device~\cite{gemma} and can offer offline capabilities so that LLMs are served even under limited network connectivity~\cite{openelm}. 

(2) \textbf{Obtaining lightweight LLMs from foundation LLMs without training from scratch}. Foundation LLMs are trained on large corpora of public data~\cite{llama} and can be fine-tuned for specific tasks~\cite{mobilevlm}. Modern foundation LLMs, such as LLaMa-3, rival classic LLMs on various benchmarks~\cite{llama3}. These LLMs can fit into a single consumer-grade GPU~\cite{llama3}. There is potential to compress foundation LLMs for use in resource-constrained environments while maintaining accuracy.

The motivation of this article is to \textit{{create small language models (SLMs) from foundation LLMs with similar quality while running on fewer resources}}. 

Research on creating SLMs is based on compression methods, such as pruning~\cite{sparsegpt, llmpruner, minitron}. 
Pruning removes an individual or a group of parameters from the model using a ranking algorithm~\cite{sparsegpt}.
There are two categories of pruning: Unstructured pruning refers to setting parameter values to zero~\cite{sparsegpt}; quality is maintained while model size is unaffected.
Structured pruning refers to removing data structures containing parameters such as attention heads~\cite{llmpruner}; this reduces model size and inference latency but at the cost of model quality.
In short, existing pruning methods do not adequately balance runtime performance and LLM quality. 

Our work, \system{}, is positioned to address these shortcomings. Existing pruning methods focus on coarse-grained pruning at the global and layer level of the LLM (further discussed in Section~\ref{bg}). They prune every LLM component uniformly. This results in removing parts of the model that are critical to quality. \system{} introduces novel fine-grained pruning of LLM projections. We leverage non-uniform pruning and apply it to different components of the LLM to selectively retain critical model parts. In addition, \system{} synergistically combines unstructured and structured pruning in LLMs for the first time to create \textbf{\textit{`composite projection pruning'}}. This pruning approach can produce compressed and resource-efficient LLMs that fit in limited memory and provide fast inference while having comparable quality to the foundation LLM. The models produced by \system{} can be deployed on any hardware platform without requiring specific hardware/software accelerators. 

Our research contributions are as follows:

(1) \textbf{Mosaic}, a novel system for compressing foundation LLMs for hardware-limited environments. \system{} is 7.19$\times$ faster in producing compressed models than existing approaches.

(2) \textbf{LLM projection pruning}, a new method that maintains quality at higher compression levels. The method determines a projection outlier distribution to prune projections non-uniformly. Projection pruning is a performance-efficient extension of fine-grained LLM pruning explored in prior work. \system{} models produced by projection pruning achieve up to 84.2\% lower perplexity and 31.4\% higher accuracy than models from uniform pruning. 

(3) \textbf{Composite projection pruning}, a new approach for LLM compression that balances the benefits of unstructured pruning to maintain quality and of structured pruning to reduce model size and inference latency across various hardware platforms. For \system{} models, up to 67\% faster inference and 68\% lower GPU memory usage compared to unstructured pruning while achieving up to 36$\times$ lower perplexity than structured pruning is noted. 

The remainder of this article is organized as follows. Section~\ref{bg} presents the background for our work. \bailey{Section~\ref{method} provides an overview of the methods underpinning \system{} and the design of the \system{} modules.} Section~\ref{ex[]} evaluates \system{} against relevant baselines. Section~\ref{relatedwork} presents related work. Section~\ref{conclusion} concludes this article.

\section{Background and Motivation}
\label{bg}
\begin{figure}[t]
  \centering
  \includegraphics[width=0.5\textwidth]{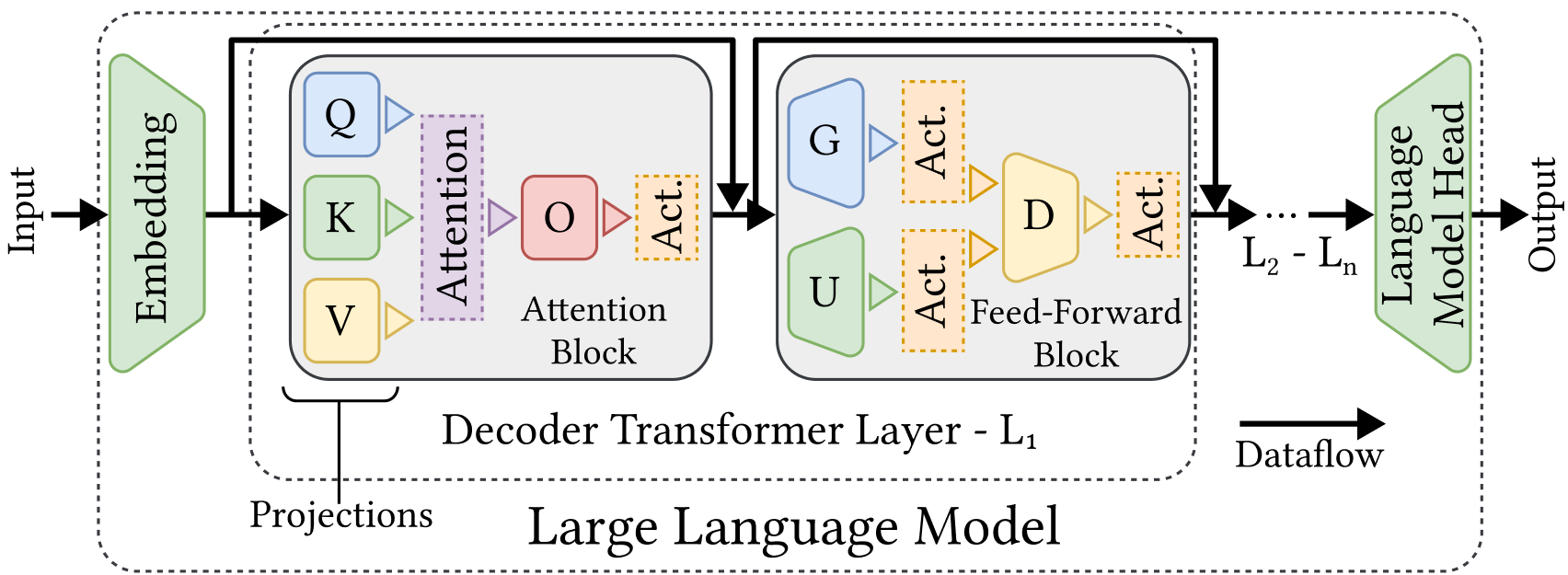}  
    \caption{Simplified overview of LLM architecture.}
    \label{fig:2n}
\end{figure}

This section considers the LLM architecture, pruning foundation LLMs to generate SLMs, the current state of LLM pruning, new opportunities in the research landscape, and the motivation for developing our approach to pruning LLMs.

\subsection{LLM Architecture}
An LLM follows the decoder transformer architecture~\cite{gpt1}.
The architecture of foundation LLMs, such as LLaMa~\cite{llama, llama3}, and fine-tuned derivatives, such as Vicuna~\cite{vicuna} is shown in Figure~\ref{fig:2n}.
It comprises the Embedding layer, the Language Model Head, and a stack of Decoder Transformer Layers, denoted as $L_1$, $L_2$, $\cdots$ $L_n$. 
The Embedding Layer transforms natural language into tokens by representing words as numerical values. The Language Model Head translates these output tokens back into natural language.

The Decoder Transformer Layers calculate attention scores for each token and use feed-forward networks to predict output tokens. Each transformer layer comprises an Attention Block and a Feed-Forward Block.
An Attention Block uses three projections — Query, Key, and Value — to calculate the relevance of each token to other tokens, generating a matrix of attention scores. The result then passes through the Output projection to the succeeding layers.
A Feed-Forward Block expands the attention scores into a larger dimension. It consists of three projections - Gate, Up, and Down. The gate controls data flow to the Up projection, which expands the dimensions, and the Down projection shrinks them.

\textbf{Projections} are the smallest units in LLMs, which contain model parameters learned during training. There are seven projections for each decoder transformer layer, which can be shortened to the set $\{Q, K, V, O, G, U, D\}$.
The \textit{parameters within the projections} are integral to the LLM. For instance, LLaMa-7B contains 7 billion parameters across 32 layers. The number of parameters determines the resources required by the LLM, which impacts model size, inference time and runtime memory use.

\begin{figure}[t]
  \centering
  \includegraphics[width=0.5\textwidth]{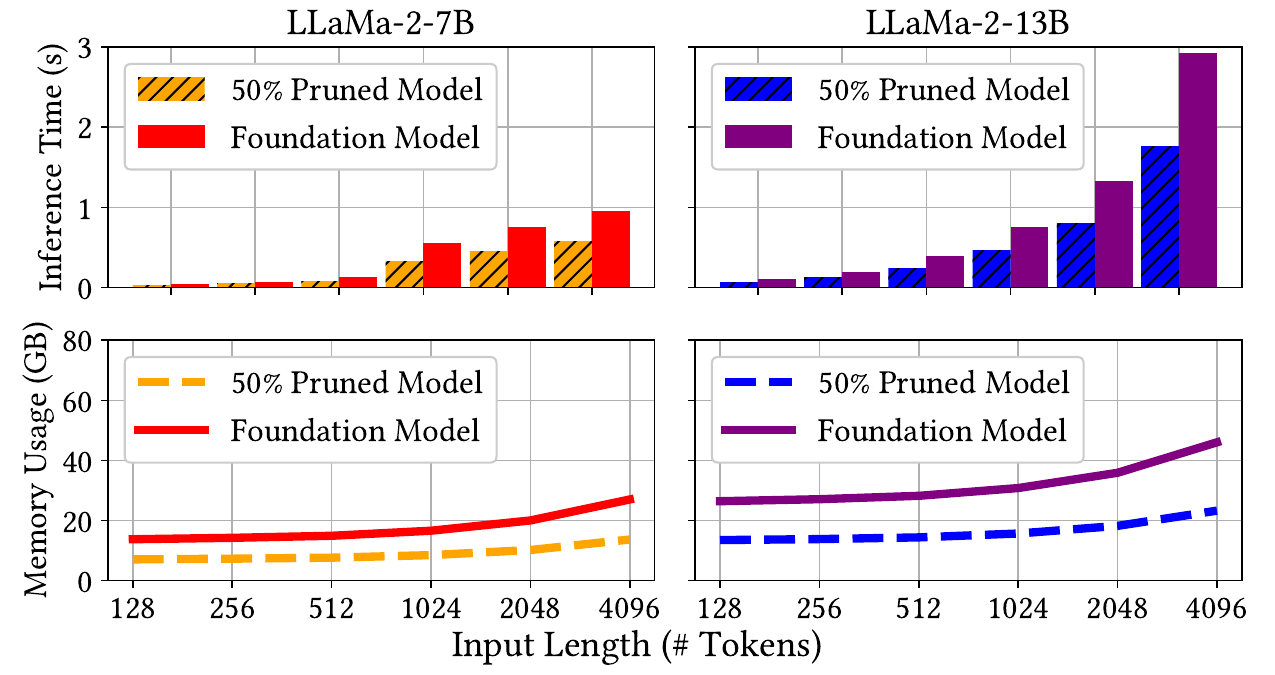}  
    \caption{GPU memory required and inference time of LLaMa-2-7B, LLaMa-2-13B, and the variants of these LLMs uniformly pruned by 50\% for varying input sizes. \bailey{Metrics were collected on an Nvidia A100 GPU using PyTorch 2.3.0 for inference.}}
    \label{fig:2}
\end{figure}

\subsection{LLM Pruning Reduces Resource Footprint}
LLMs require large volumes of input data, called tokens, to understand the context of natural language sentences~\cite{context}. Each token ($t$) interacts with every other token when calculating attention scores, resulting in a quadratic increase in memory use ($t^2$)~\cite{transformers}. In Figure~\ref{fig:2}, LLaMa-2-13B requires nearly 20 GB more memory at 4096 tokens than 128 tokens, increasing memory overheads by 77\% over the original model size. More tokens also increase inference time since the number of attention score calculations increases~\cite{flashattention}. LLaMa-2-13B inference time increases 30$\times$ from 0.1s to nearly 3s.

\textbf{LLM pruning} reduces resource demand on hardware-limited devices by removing LLM parameters. Removing parameters has a \textit{two-fold benefit}: (1) reduced model size and (2) reduced attention and activation matrix memory sizes during inference. In Figure~\ref{fig:2}, LLM pruning of LLaMa-2-7B and LLaMa-2-13B by 50\% reduces the parameter count to 3.5B and 6.5B, respectively. The dashed lines and the hatch-filled bars represent the reduced resource usage of pruned LLMs compared to foundation LLMs. Pruned LLMs are 2$\times$ smaller and 40\% faster. Foundation LLMs are available in many fixed sizes, such as LLaMa-2 with a 7B, 13B, 34B and 70B variant~\cite{llama2}. While larger variants are targeted for multi-GPU deployments, existing LLM pruning methods are positioned to compress the smaller variants for GPUs that do not have the compute and memory resources to run them~\cite{wanda, owl}.

\subsection{Opportunities in LLM Pruning}
\textbf{Limitations of Current LLM Pruning Methods:}
LLM pruning requires ranking the importance of each parameter in relation to the overall model quality~\cite{obd}. The lowest-ranking parameters are removed based on a target pruning percentage~\cite{pruning_survey}. For example, the lowest ranking 30\% of parameters is removed if the pruning target is 30\% compression. 

To calculate the importance of each parameter, hundreds of data samples pass through the LLM to activate each parameter and then rank them~\cite{sparsegpt, wanda}. However, consider Figure~\ref{fig:2}, where memory usage can go above the capacity of a single consumer GPU. Current LLM pruning methods rank parameters one layer at a time to reduce memory usage~\cite{sparsegpt}. This approach results in parameters ranked by importance within individual layers rather than globally. Consequently, the pruned LLMs are of lower quality, and each layer is pruned uniformly~\cite{owl}. If LLM pruning evaluates the importance of every parameter globally during ranking, then parameters could be pruned based on their importance relative to the entire LLM rather than their importance within individual layers. This allows non-uniform pruning across the model - targeting and removing redundant parameters while preserving crucial parameters required to maintain model quality.

\begin{figure}[t]
  \centering
  \includegraphics[width=0.5\textwidth]{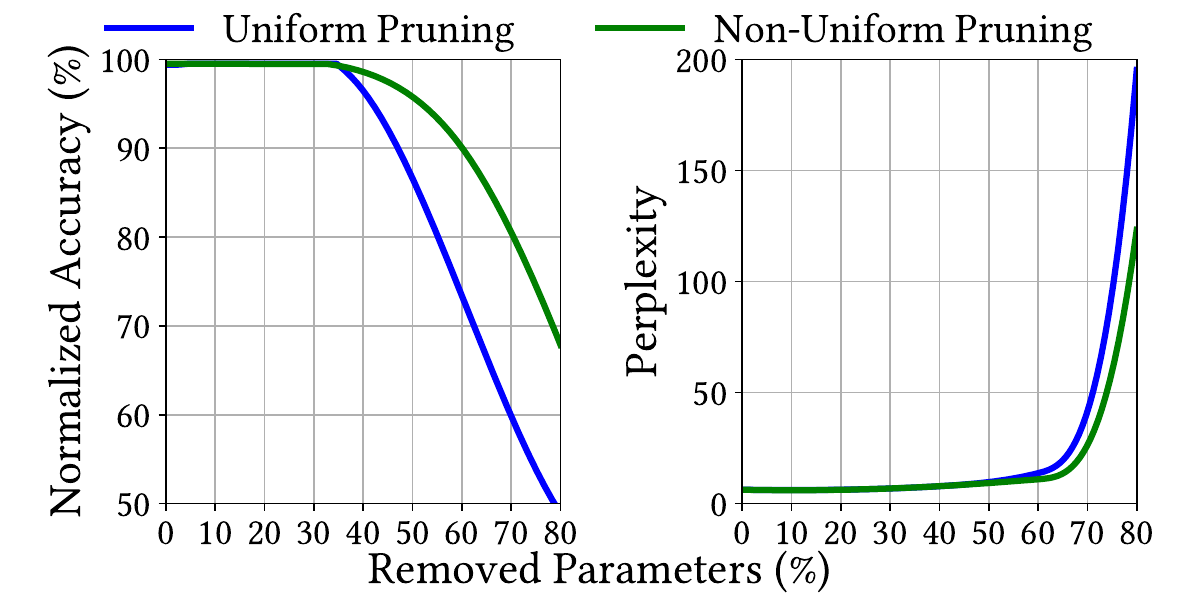}  
    \caption{Normalized accuracy (higher is better) and perplexity (lower is better) of LLaMa-3-8B as parameters are removed via uniform and non-uniform pruning.}
    \label{fig:4}
\end{figure}

In \textbf{uniform pruning}, every component (i.e. layer, block, and projection) of the LLM is pruned by the same amount, whereas in \textbf{non-uniform pruning}, some components are pruned more than others.
\bailey{Figure~\ref{fig:4} shows the accuracy and perplexity achieved by uniform and non-uniform pruning on language benchmarks for the foundation LLM, LLaMa-3-8B.} As more parameters are removed, non-uniform pruning achieves higher accuracy and lower perplexity than uniform pruning since non-uniform pruning can more accurately rank and prune LLMs. For the same accuracy loss, non-uniform pruning can eliminate up to 25\% more parameters compared to uniform pruning, removing 70\% versus 55\% of the parameters, respectively. When 70\% of parameters are removed, non-uniform pruning achieves a 40\% lower perplexity of 25 compared to uniform pruning that has 40. 

\begin{tcolorbox}[
    width=0.48\textwidth,
    colframe=black!50!black,
    colback=gray!5,boxsep=3pt,
    boxrule=0.5pt, left=3pt,right=3pt,top=1pt,bottom=1pt]
\textbf{Opportunity 1:} Non-uniform pruning improves accuracy and perplexity compared to uniform pruning.
\end{tcolorbox}

\textbf{Defining Projection Pruning:}
We first consider global and layer/block pruning before defining projection pruning. 

\textit{Global Pruning} is synonymous with uniform pruning. It was introduced to prune LLMs with lower GPU memory overheads~\cite{sparsegpt}. Pruning occurs globally for a fixed percentage~\cite{wanda}. Each layer, block, and projection within the model removes the same percentage of parameters without accounting for their importance between different components. \bailey{While projections are pruned, they are not evaluated individually for importance; instead, they are uniformly pruned by applying the same percentage across all projections.}

\bailey{\textit{Layer/Block-wise Pruning} is a more fine-grained pruning approach (we refer to as quasi-non-uniform layer/block pruning)} in which each layer/block has a different pruning percentage~\cite{pruning_survey}. For convolutional neural networks (CNNs), pruning layers by different percentages allows important parameters in sensitive layers to be untouched while aggressively pruning redundant layers~\cite{reconvene}. This leads to better model accuracy at higher sparsities and extends to LLMs~\cite{owl}. However, each projection within each layer/block is pruned to the same percentage, which does not account for the importance of parameters between the seven projections within each layer.

In this article, we define \textbf{Projection Pruning} as the finest-grained non-uniform pruning method of the projections within an LLM. Each projection is pruned to a specific percentage based on how important each parameter is against other projections of the same category. For example, all query projection parameters in a given transformer layer are ranked against all query projections across all layers. This granularity of pruning is unexplored for LLMs. Our work explores projection pruning for the first time. 

\begin{tcolorbox}[
    width=0.48\textwidth,
    colframe=black!50!black,
colback=gray!5,boxsep=3pt, boxrule=0.5pt, left=3pt,right=3pt,top=1pt,bottom=1pt]
\textbf{Opportunity 2:} The proposed projection pruning, a non-uniform pruning method, offers more control over the parameters removed when pruning.
\end{tcolorbox}

\textbf{Combining Unstructured and Structured Pruning:} 
Uniformity determines where and how much pruning occurs within the model, while unstructured and structured pruning are categories that specify how parameters are removed. 

\textit{Unstructured pruning (UP)}~\cite{sparsegpt,wanda} sets parameter weights to zero, creating model sparseness; although zeroed parameters no longer contribute to the model, the size of the model remains unchanged. Sparse models retain model quality but often require vendor-specific GPUs and acceleration libraries for limited speedup gains - for example, only 1.24$\times$ speedup at 50\% sparsity for LLaMA-7B~\cite{wanda}.

\textit{Structured pruning (SP)}~\cite{llmpruner} removes entire data structures containing groups of parameters, thus significantly reducing the model size and inference latency. However, SP affects model quality, which decreases more rapidly for higher sparsities than UP. Structured pruning methods target resource-constrained devices that cannot run the original full-sized model. Typically, these devices do not have GPUs or support the libraries for sparse model inference.

In this work, we leverage the benefits of unstructured and structured pruning (i.e., retaining model quality while reducing model size and inference latency) within projection pruning. 
This combination is referred to as \textbf{Composite Projection Pruning}, which is unexplored for LLMs. Projection pruning has been unexamined due to the challenges in accelerating non-uniformly pruned projections~\cite{owl}. The integration of unstructured and structured pruning allows for accelerating these models, even when specialized accelerators are unavailable~\cite{reconvene, ECCLES202443}, thereby addressing the problem of non-uniformity. Previous composite pruning methods~\cite{reconvene, ECCLES202443} have focused on convolutional neural networks (CNNs), wherein each layer comprises a single type of component, for example, a collection of convolution filters or a fully connected linear layer. In contrast, layers of LLMs consist of multiple blocks featuring diverse projections of varying dimensions and purposes, which has not been investigated within the framework of composite pruning.

\begin{tcolorbox}[
    width=0.48\textwidth,
    colframe=black!50!black,
    colback=gray!5,
colback=gray!5,boxsep=3pt,boxrule=0.5pt,left=3pt,right=3pt,top=1pt,bottom=1pt]
\textbf{Opportunity 3:} Combining unstructured and structured pruning for projections, referred to as \textit{composite projection pruning}, retains model quality while reducing model size and inference latency.
\end{tcolorbox}

\bailey{Global/Layer/Projection pruning alone refers to the granularity of the pruning method, which can be applied in an unstructured or structured manner. Therefore, the potential dimensions of projection pruning that are explored in this manuscript include unstructured, structured, and composite projection pruning.}

\subsection{Leveraging the Opportunities}
We present \system{}, a system that leverages the above opportunities and creates a novel approach for deriving and deploying compressed LLMs on resource-constrained hardware. \system{} models bridge the performance and quality gap between UP and SP, allowing for flexible deployment of existing foundation LLMs to multiple target hardware platforms. While composite projection pruning is unique to \system{}, the system can create pruned models using any of the three categories of model pruning depending on the available resources for the target platform. For example, a \system{} LLM using UP will suit a GPU-rich environment to achieve high model quality. A \system{} model derived using SP makes LLM inference possible at a minimal accuracy loss in environments that have limited resources. In addition, \system{} models derived using composite projection pruning may be suited for weak GPUs and reducing the overall memory footprint.

\section{Design of \system{}}
\label{method}
\system{} adopts composite projection pruning on pre-trained foundation LLMs to create SLMs for a target device. \system{} builds on the groundwork laid by global, layer, and block pruning while combining unstructured and structured pruning for the first time on LLMs and applies it for projections to achieve true non-uniform pruning. This section explores the proposed composite projection pruning method. When \system{} is employed solely for unstructured or structured pruning, it utilizes the implementations outlined in Section~\ref{pm}.

\subsection{Pruning LLMs Across Projections}
\label{mosaic:method}

In the simplest case, global/uniform pruning applies the same pruning target $p$ to every component of the LLM where $p \in [0,1)$. For example, $p=0.3$ prunes 30\% of the model parameters. For non-uniform pruning, a pruning \textit{target} must be calculated for each component. This subsection presents the underlying concepts of projection pruning. Firstly, a ranking method is proposed to select a varying set of pruning targets for each layer from the initial $p$ value to achieve non-uniform projection pruning. Secondly, the Projection Outlier Distribution (POD) is defined to achieve projection pruning.

\subsubsection*{From Layer to Projection Pruning} 
Layer pruning creates a pruning target for each layer, which overall averages to approximately $p$. For a LLM with $N$ layers, 

\begin{equation}
    p \approx \frac{\sum_{n=1}^N p_n}{N}
    \label{eq:1-}
\end{equation}

where $p_n$ is the pruning target of the $n^{th}$ layer. In layer pruning, $p_n$ is applied to all projections within the layer. Each layer has a unique pruning target, but the projections within each layer are pruned by the same amount (Figure~\ref{fig:2n}).

\textit{Projection pruning} extends layer pruning by determining a pruning target for each projection. For an LLM with $M$ projections, 
\begin{equation}
   p_n \approx \frac{\sum_{m=1}^M p_{n,m}}{M}
   \label{eq:2-}
\end{equation}
where $p_{n,m}$ is the pruning target of the $m^{th}$ projection in the $n^{th}$ layer. For example, $p_{1,1}$ is the pruning target for the Query projection in the first layer, where $p_{2,2}$ is the Key projection in the second layer (Figure~\ref{fig:2n}). Working backwards, using Equation~\ref{eq:1-}, $p$ can be found for the entire LLM.

The number of pruning targets increases with the granularity of the pruning method. While global pruning has one pruning target, layer pruning has $N$ pruning targets, and projection pruning has $N \cdot M$. For example, LLaMa-7B has $N=32$ and $M=7$ or 32 and 224 pruning targets for layer and projection pruning, respectively.

\subsubsection*{Projection Outlier Distribution} 
In global pruning, $p$ is typically defined by the system user. However, pruning targets must be derived from an initial $p$ value for layer and projection pruning. Existing layer pruning methods use Layer Outlier Distribution (LOD)~\cite{owl} to calculate $p_n$ for every layer. LOD creates a set of ratios based on how many outlier parameters are in each layer. The ratios are then scaled and normalized by $p$ to calculate $p_n$ for each layer. 

Parameter outliers of a layer are defined as the set of parameters with a \textit{weight metric} greater than the average for the layer. Existing research uses the weight metric $\omega_n$ for layer $n$~\cite{wanda, owl}:
\begin{equation}
   \omega_n = \|A_n\|_2 \cdot |\theta_n|
   \label{eq:3}
\end{equation} 
where $\|A_n\|_2$ is the $l_2$ norm of the activations of layer $n$ and $|\theta_n|$ is the magnitude of parameter weights in layer $n$.
LOD determines that for each parameter $i$ in $\theta_n$ are outliers in $n$ if:
\begin{equation}
   IsLayerOutlier(\theta_n^i) = \omega_n^i > \alpha \cdot \overline{\omega_n}
   \label{eq:4}
\end{equation}
is true, where the weight metric $\omega_n^i$ of a parameter $i$ is greater than the threshold of a constant $\alpha$, typically set to five or greater~\cite{owl}, multiplied by the average weight metric for the layer. The number of outliers per layer is then scaled into a ratio contrasted by the other layers to determine $p_n$ for each layer. Layers with more outliers are important to LLM model accuracy~\cite{wanda}; therefore, they are pruned less than layers with fewer outliers. Important layers are assigned smaller $p_n$ values, while less critical layers have larger $p_n$ values to achieve the overall pruning target $p$ (Equation~\ref{eq:1-}).

To extend LOD for projection pruning and determining a set of $p_{n,m}$ values, \system{} defines \textit{Projection Outlier Distribution (POD)} by making the following fundamental changes: (1) the weight metric is calculated at the projection level and (2) outlier parameters are compared within the same projection instead of across the layer. 

To achieve (1), Equation~\ref{eq:3} is modified as follows:
\begin{equation}
   \omega_{n,m} = \|A_n\|_2 \cdot |\theta_{n,m}|
   \label{eq:5}
\end{equation}
where the $l_2$ norm term remains the same; however, a weight metric for each projection $m$ is calculated
for each layer $n$ using only the magnitude of weights in each projection $m$. This allows Equation~\ref{eq:6} to find projection outliers as the weight metric of each parameter is restricted to the set of parameters in that particular projection $\omega_{n,m}$ which achieves (2). We define POD using projection outliers with 
\begin{equation}
   IsProjectionOutlier(\theta_{n,m}^i) = \omega_{n,m}^i > \alpha \cdot \overline{\omega_{n,m}}
   \label{eq:6}
\end{equation}

\subsection{Composite Projection Pruning}
In existing literature, structured pruning accelerates sparsely trained CNNs~\cite{ECCLES202443}, such as VGG-16~\cite{vgg}. This is achieved by first training a sparse model that is pruned using unstructured pruning. After training, channels that have entire zero values (as an outcome of unstructured pruning) are removed via structured pruning. This work defines composite projection pruning for the first time for LLMs, where unstructured and structured pruning are simultaneously employed across projections. This contrasts with prior work using unstructured and structured pruning in a sequence for CNNs. 

\begin{figure}[t]
    \centering
    \includegraphics[width=0.45\textwidth]{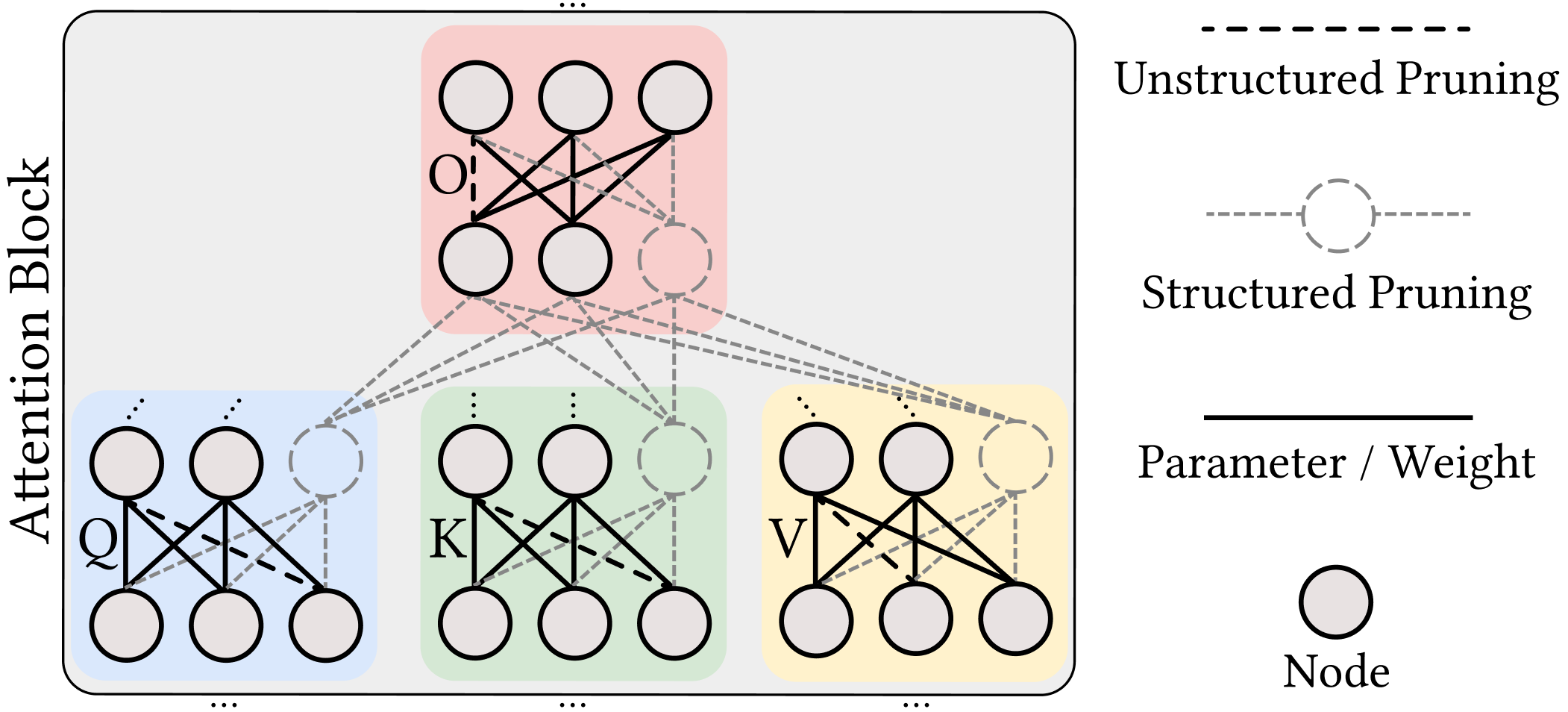}  
    \caption{Example of combining unstructured and structured pruning in composite projection pruning.}
    \label{fig:composite}
\end{figure}

Figure~\ref{fig:composite} shows that for an LLM attention block, parameters in each projection are individually pruned via unstructured pruning based on POD (Section~\ref{mosaic:method}) and simultaneously projection nodes between projections are pruned as a group via structured pruning. This combination is referred to as \textit{composite projection pruning} and to derive SLMs from LLMs that are of high quality with low resource overheads. Section~\ref{e3} will explore the benefits of composite projection pruning across a range of hardware. 

Next, the modules of \system{} are considered. It takes a foundation LLM as input, and using POD, determines $p_{n,m}$ for each projection to prune the LLM by $p$ using unstructured, structured, and composite projection pruning.
\begin{figure}[t]
    \centering
    \includegraphics[width=0.44\textwidth]{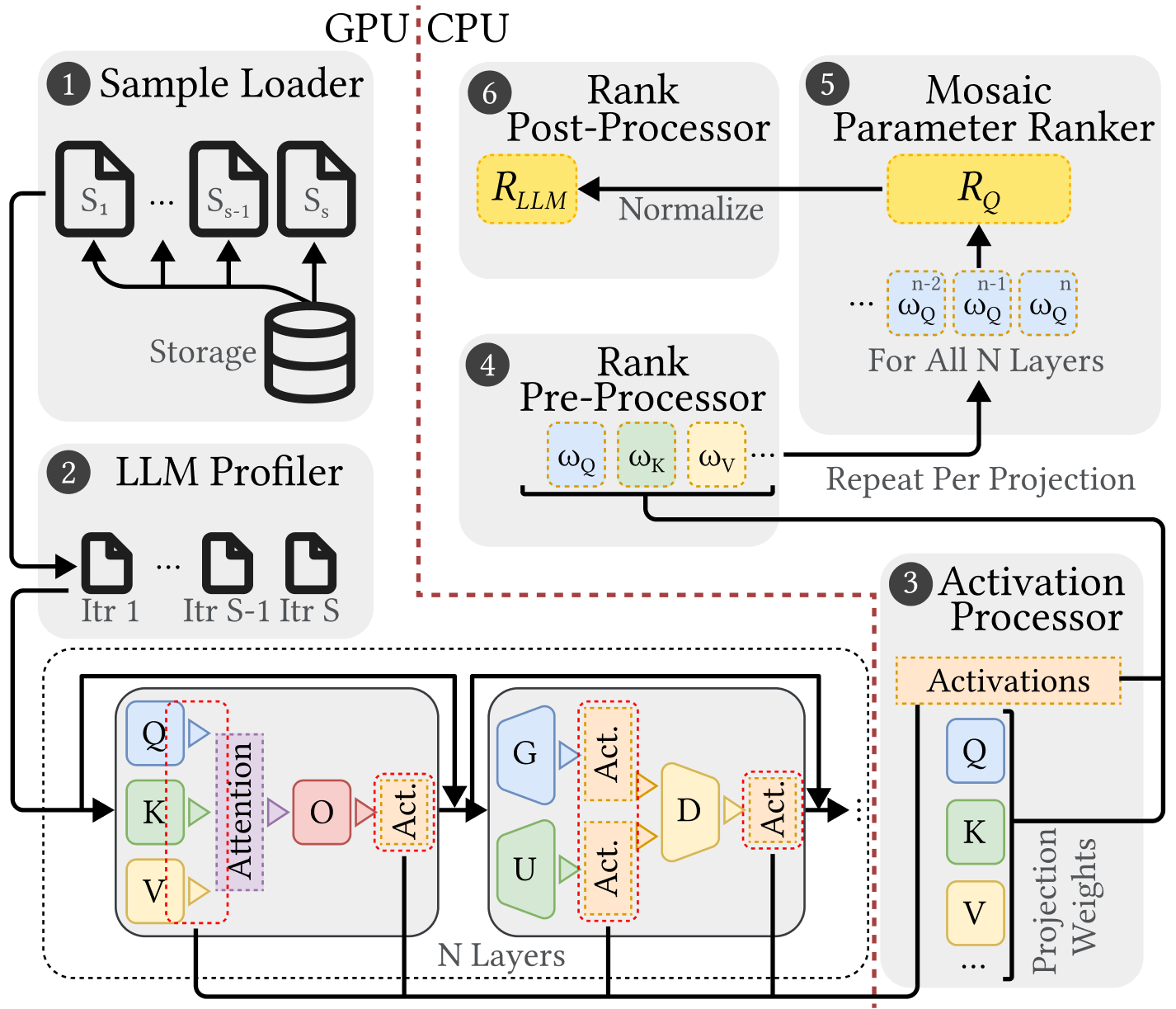}  
    \caption{Overview of the \system{} Parameter Ranking Controller. A global rank ($R_{LLM}$) captures the importance of each projection in every layer which is calculated using a combination of activations and the projection weights.}
    \label{fig:5}
\end{figure}

\subsection{\system{} System}

\begin{figure}[t]
  \centering
  \hspace*{-10pt}
  \includegraphics[width=0.52\textwidth]{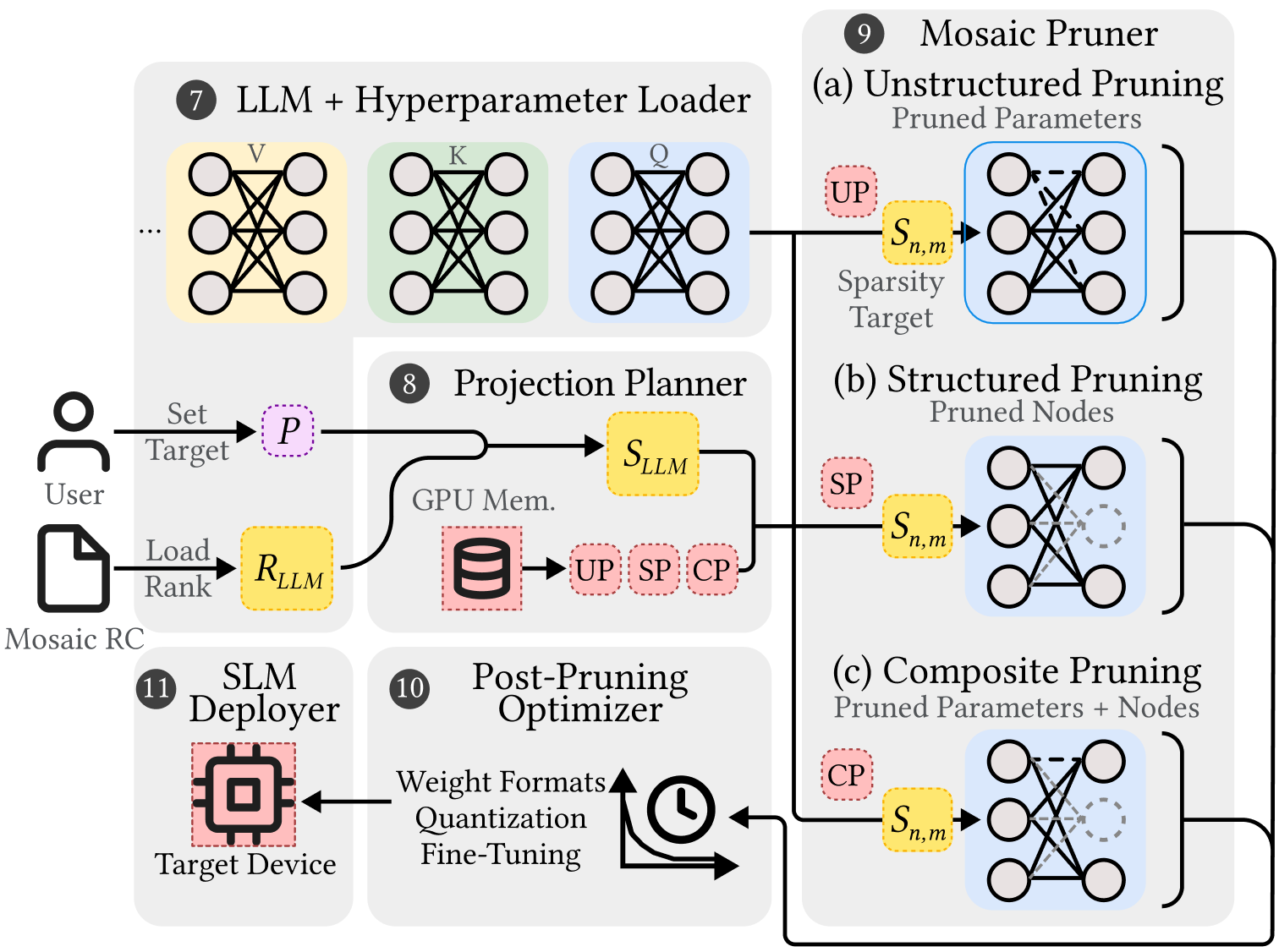}  
    \caption{Overview of the \system{} Parameter Pruning Controller. The $R_{LLM}$ from the \system{} RC acts as a look-up table to scale the pruning target to a sparsity target per projection.}
    \label{fig:6}
\end{figure}

\system{} consists of two modules - Parameter Ranking Controller (Figure~\ref{fig:5}) and Parameter Pruning Controller (Figure~\ref{fig:6}) that are run sequentially to deploy an LLM on a target hardware platform. The modules are detailed below.

\textbf{Parameter Ranking Controller (RC)} profiles the LLM to create a \textit{global rank} ($R_{LLM}$) that represents the importance of each projection by identifying projection parameter outliers. Each LLM is profiled once to reuse the global rank for any pruning level $p$ across different pruned LLM variants. As a result, the initial overhead (Section~\ref{e5}) in creating the global rank is offset by faster inference achieved by the pruned LLMs.

The RC comprises six components as shown in Figure~\ref{fig:5}. After preloading the LLM into memory, the {\color{white}\cir{1}}~\textit{Sample Loader} moves a small calibration set of tokens into memory (128 samples $\times$ 2048 tokens $\times$ $\sim$4 bytes per token = $\sim$1 KB). 

The {\color{white}\cir{2}}~\textit{LLM Profiler} infers the LLM model with each sample iteratively. {\color{white}\cir{1}} and {\color{white}\cir{2}} are executed on the GPU. The {\color{white}\cir{3}}~\textit{Activation Processor} hooks into the activation function for each projection for every sample and transfers the resulting activations to the CPU. The activations are a proxy for the $l_2$ norm term in Equation~\ref{eq:5}. 

The {\color{white}\cir{4}}~\textit{Rank Pre-Processor} calculates the weight metric (Equation~\ref{eq:5}) for each projection using the captured activations and projection weights. The {\color{white}\cir{5}}~\textit{Mosaic Parameter Ranker} calculates the POD (Equation~\ref{eq:6}) for each projection by comparing each parameter weight metric against the average weight metric, thus identifying outliers. The projections with the highest number of outliers are adjusted accordingly.

Finally, the {\color{white}\cir{6}}~\textit{Rank Post-Processor} normalizes all projection ranks into a global rank, which captures the importance of each projection against every other projection in the LLM. The global rank is the output of the RC module and serves as the input to the next module.

\textbf{Parameter Pruning Controller (PC)} prunes the LLM for deployment as shown in Figure~\ref{fig:6} using five components executed on the CPU. 

The {\color{white}\cir{7}}~\textit{LLM + Hyperparameter Loader} loads the LLM into GPU memory and the global rank from the RC and a pruning target defined by the user where $p \in [0,1)$. 

The {\color{white}\cir{8}}~\textit{Projection Planner} scales the global rank by the pruning target, creating a \textit{sparsity target} for each projection in the LLM. The average sparsity target across all projections equals $p$ (Equation~\ref{eq:1-} and Equation~\ref{eq:2-}). In addition, the available GPU memory of the target deployment platform is used to determine the pruning category. 

The {\color{white}\cir{9}}~\textit{Mosaic Pruner} prunes each projection by the sparsity target using the identified pruning category, which considers factors such as whether the target platform has enough memory to load the LLM after pruning. Three pruning categories are considered as follows: 

\textit{(a) Unstructured Projection Pruning} suitable for cloud-tier hardware such as server-grade GPUs where memory capacity and bandwidth are not necessarily a bottleneck, and the availability of sparsity accelerators such as NVIDIA CUTLASS can leverage the sparsity from unstructured pruning. 

\textit{(b) Structured Projection Pruning} is chosen for low-end edge devices where GPUs may be unavailable and are typically memory-limited. 

\textit{(c) Composite Projection Pruning} combines (a) and (b) categories and prunes in that order. Composite pruning is reserved for platforms with mobile or older-generation GPUs.

The LLM is then pruned using the determined pruning category, where each projection is reduced by the sparsity target, creating an overall pruned LLM. 

The {\color{white}\cir{10}}~\textit{Post-Pruning Optimizer} prepares the pruned LLM for deployment, including steps to improve downstream task performance such as fine-tuning low-rank adaptors (LoRA)~\cite{lora}, further compression such as quantization~\cite{gptq}, or converting the model weights into different inference formats for specific accelerators. For example, ONNX weights for DeepSparse CPU accelerators~\cite{deepsparse}.

The {\color{white}\cir{11}}~\textit{SLM Deployer} deploys the pruned LLM, which is a small language model (SLM)~\cite{phi}, to the target device.

\begin{algorithm}[t]
\setstretch{1.15}
\caption{\system{} Parameter Ranking Controller (RC)}
\label{alg:1}
\begin{algorithmic}[1]
\REQUIRE Calibration samples $S$, Foundation large language model $LLM$, Projection Outlier Constant $\alpha$
\ENSURE Global rank $R_{LLM}$
\STATE $S$.GPU();  $LLM$.GPU(); \small{ // Move $S$ and $LLM$ to GPU mem.}
\STATE $N$ $\leftarrow$ \texttt{length}($LLM.layers$\texttt{)}; \small{ // \# LLM Layers}
\STATE $M$ $\leftarrow$ \texttt{length(}$LLM.layers[0].projs$\texttt{)}; \small{ // \# LLM Projections}
\STATE $R_{LLM}$ $\leftarrow$ [$N$][$M$]; \small{ // Empty list of lists of global rank for each projection for each layer}

\FOR{$n$ in $LLM.layers$}
    \STATE $A$ $\leftarrow$ $[]$; \small{ // Empty list of layer activations}
    \FOR{$s$ in $S$}
        \STATE $A[s] = LLM.layers[n]$\texttt{.infer(}$s$\texttt{)};\scriptsize{ // Capture activations of $n$}
    \ENDFOR
    \FOR{$m$ in $LLM.layers[n].projs$}
        \STATE $\omega_{n,m}$ $\leftarrow \|A_n\|_2 \cdot |\theta_{n,m}|$; \small{ // Weight metric, Equation 5}
        \STATE $\omega_{n,m} \leftarrow $ \texttt{cat(flatten(}$\omega_{n,m}$\texttt{))}; \small{ // Reshape $\omega_{n,m}$}
        \STATE $C_{n,m}$ $\leftarrow$ \texttt{numel(}$m$\texttt{)}; \small{ // \# Parameters in $m$}
        \STATE $O_{n,m}$ $\leftarrow$ $\sum_{i=1}^{C_{n,m}}1(\omega_{n,m}^i > \alpha \cdot \overline{\omega_{n,m}})$; \small{ // \# Outliers, Equation 6}
        \STATE $R_{n,m}$ $\leftarrow \frac{O_{n,m}}{C_{n,m}} \cdot 100;$ \small{ // Projection rank of $m$}
        \STATE $R_{LLM}[n][m] \leftarrow R_{n,m}$; \small{ // Update global rank $R_{LLM}$}
    \ENDFOR
\ENDFOR
\STATE \textbf{return} \texttt{normalize($R_{LLM}$)}; 
\end{algorithmic}
\end{algorithm}

\subsection{Implementation}
\label{mosaic:imp}
\system{} is implemented using Python 3.8.10, PyTorch~2.3.0, Transformers 4.43.1, and CUDA 11.7. 

\bailey{\textbf{Algorithm~\ref{alg:1}} shows the procedure of the \system{} Parameter Ranking Controller (Figure~\ref{fig:5}).}
The RC generates the global rank using Equation~\ref{eq:5} and Equation~\ref{eq:6} to generate pruning targets for projections in the PC.
First, the calibration samples $S$, the $LLM$ to prune, and a constant $\alpha$ for Equation~\ref{eq:6} are loaded into system memory. The samples and LLM are then moved to the GPU memory (Line~1). Metadata, such as the number of layers $N$ and projections $M$, are calculated by measuring the LLM dimensions (Lines~2-3). Lastly, an empty list of lists to store the global rank is initialized using $N$ and $M$ (Line~4). The CPU portion of the algorithm is then activated layer by layer as the samples are passed through the LLM (Lines~6-9). $A$ captures the activations (Line~8). For each projection, the weight metric is calculated by multiplying the $l_2$ norm of the activations by the magnitude of the projection weights (Equation~\ref{eq:5}, Line~11). The weight metric vector is flattened and concatenated across samples (Line~12). The mean of the weight metric is used with $\alpha$ (Line~14) to calculate the ratio of projection outliers (Equation~\ref{eq:6}, Line~15) by dividing the number of outliers by the total parameters in the projection (Line~13). The global rank is then updated for the projection in the current layer (Line~16). The algorithm calculates the global rank for every projection across every layer $R_{LLM}$. Finally, the normalized global rank is returned (Line~19).
 
\section{Experiments}
\label{ex[]}
\newcommand{\rnum}{\stepcounter{rownum}\therownum}

\begin{table}[t]
\caption{Hardware platforms used in the evaluation.}
\label{tab:hardware}
\centering
\footnotesize
\begin{tabular}{crr}
\hline
\tiny{Platform} & \footnotesize{CPU} \scriptsize{(Arch.)} & \footnotesize{GPU (Per GPU: Mem., Bandwidth)} \\ \hline
P\rnum & \footnotesize{AMD EPYC 7713} \scriptsize{(x86)}    & \footnotesize{2$\times$ Nvidia A100} \scriptsize{(80 GB, 1935 GB/s)} \\
P\rnum & \footnotesize{AMD EPYC 7713P} \scriptsize{(x86)}   & \footnotesize{2$\times$ Nvidia A6000} \scriptsize{(48 GB, 768 GB/s)}  \\
P\rnum & \footnotesize{Intel i9-13900KS} \scriptsize{(x86)} & \footnotesize{Nvidia RTX 3080} \scriptsize{(10 GB, 760 GB/s)}  \\
P\rnum & \footnotesize{Cortex-A78AE} \scriptsize{(ARM)}     & \footnotesize{Nvidia AGX Orin} \scriptsize{(64 GB*, 205 GB/s)}  \\
P\rnum & \footnotesize{Cortex-A76} \scriptsize{(ARM)}       & \scriptsize{Broadcom VideoCore VII} \scriptsize{(4 GB$\dagger$, 15 GB/s)}   \\ \hline
\end{tabular}
\smallskip
\footnotesize{*Combined system and GPU memory capacity\\\vspace{-3pt}$\dagger$Maximum GPU mem. assigned from a shared 8 GB system mem. pool.}
\vspace{-12pt}
\end{table}

\begin{table*}[t]
\caption{LLM architecture and training details. The model size is for parameters in FP16 half precision. Attention Dimensions and Feed-Forward Dimensions are the number of channels in the innermost dimension of each projection.}
\label{tab:models}
\centering
\small
\begin{tabular}{lrccccrr}
\hline
Model  & 
Params. & 
Layers &
Attention Dim. &
Feed-Forward Dim. &
Model Size &
Training Size &
Context Len. \\ \hline
LLaMa-3.1-8B~\cite{llama31} & 8.03B & 32 & 4,096 & 14,336 & 16.07 GB & \textgreater 15T & 128K \\
LLaMa-3-8B~\cite{llama3}   & 8.03B & 32 & 4,096 & 14,336 & 16.07 GB & $\gtrsim$15T & 8K \\
LLaMa-2-13B~\cite{llama2}   & 13.02B & 40 & 5,120 & 13,824 & 26.03 GB & 2T & 4K \\
LLaMa-7B~\cite{llama}     & 6.74B & 32 & 4,096 & 11,008 & 13.48 GB & 1.4T & 2K \\
Vicuna-7B v1.5~\cite{vicuna} & 6.74B & 32 & 4,096 & 11,008 & 13.48 GB & 2T + 0.37B* & 4K                \\
\hline
\end{tabular}
\smallskip
\footnotesize{*Foundation model training size plus fine-tuning dataset size.} \vspace{-10pt}
\end{table*}

This section presents the experimental setup in Section~\ref{exp:setup} and the results from evaluation in Section~\ref{subsec:results}.  

\subsection{Experimental Setup}
\label{exp:setup}

\subsubsection{Testbed} Five hardware platforms as shown in Table~\ref{tab:hardware} are used as the evaluation testbed: (P1) A 2$\times$ A100 GPU system akin to a Google Cloud \texttt{a2-ultragpu-2g} instance running Ubuntu 20.04.06 LTS with kernel version 5.4.0-193-generic is used for all experiments unless stated otherwise. (P2) A 2$\times$ A6000 GPU system akin to cloud GPU servers with lower GPU memory/bandwidth than P1. (P3) A consumer desktop GPU. (P4) A small form factor SoC GPU. (P5) A Raspberry Pi 5 with limited GPU memory and bandwidth.

\subsubsection{Evaluated Models} Five LLMs with varying characteristics shown in Table~\ref{tab:models} are considered. Each model has seven projections per layer. All pruning methods are evaluated on each model for quality and performance metrics for a range of sparsity values.
The LLMs have the following characteristics: 

(1) \textit{Parameter Count -} four different sizes: 6.74 billion, 7.37 billion, 8.03 billion, and 13.02 billion parameters. 

(2) \textit{Model Depth -} model depth of 32 and 40 layers.

(3) \textit{Block Parameter Distribution -} ranging from 1:2.7 to 1:3.5 attention block to feed-forward parameter ratio ($\sfrac{Attention\;Dim.}{Feed-Forward\;Dim.}$).

(4) \textit{Extent of Training -} each LLM is trained with different data sizes, ranging from 1.4 trillion to over 15 trillion tokens.

(5) \textit{Fine-tuned Parameters -} Vicuna-7B v1.5 is a fine-tuned LLM for conversational tasks, whereas LLaMa models are foundation models without fine-tuning.

(6) \textit{Context Length -} from 2K to 128K token context length.

\subsubsection{Pruning Methods}
\label{pm}
Three \textit{pruning uniformity methods} that use SparseGPT~\cite{sparsegpt} to prune the lowest ranking parameters using the inverse Hessian matrix and a subsequent weight update in a one-shot manner are considered:

1) \textbf{Global Pruning -} each component is pruned uniformly by $p$ as presented in Section~\ref{mosaic:method}. 

2) \textbf{Layer Pruning -} implemented by OWL~\cite{owl} and uses the Wanda~\cite{wanda} weight metric (Equation~\ref{eq:3}) to identify LOD (Equation~\ref{eq:4}) across layers. Using LOD and $p$; OWL derives $p_n$ for each layer (Equation~\ref{eq:1-}). Each layer is pruned by $p_n$.

3) \textbf{Projection Pruning -} implemented by \system{} that uses the weight metric (Equation~\ref{eq:5}) to identify POD presented in Equation~\ref{eq:6} across projections. Using POD and $p$, \system{} derives $p_{n,m}$ for each projection (Equation~\ref{eq:2-}). Each projection is pruned by $p_{n,m}$.

Three \textit{pruning category methods} are implemented as: 

1) \textbf{Unstructured Projection Pruning -} pruned parameters are masked by setting their weights to zero. A zeroed parameter is considered pruned.

2) \textbf{Structured Projection Pruning -} parameters are pruned by removing attention and feed-forward heads and channels using LLM-Pruner~\cite{llmpruner} as illustrated in Figure~\ref{fig:composite}.

3) \textbf{Composite Projection Pruning -} \bailey{\system{} first applies unstructured pruning to individual parameters, followed by structured pruning to remove the lowest-magnitude attention and feed-forward heads. This approach is analogous to prior composite pruning methods developed for CNNs~\cite{ECCLES202443}.}

\begin{table}[t]
\caption{Datasets used in the evaluation. Batch size (BS) is fixed per dataset.}
\label{tab:datasets}
\centering
\small
\begin{tabular}{clcr}
\hline
Task & \multicolumn{1}{l}{Dataset} & Metric & BS \\\hline
\multirow{7}{*}{\begin{tabular}[c]{@{}c@{}}Common-sense\\ Reasoning\\ Accuracy\end{tabular}} 
& ARC-e~\cite{arc} & Norm. Accuracy & \multirow{7}{*}{32*}\\
& ARC-c~\cite{arc}  & Norm. Accuracy &\\
& BoolQ~\cite{boolq} & Accuracy &\\
& HellaSwag~\cite{hellaswag} & Norm. Accuracy &\\
& OBQA~\cite{obqa} & Norm. Accuracy &\\
& RTE~\cite{rte} & Accuracy &\\
& WinoGrande~\cite{winogrande} & Accuracy &\\ \hline
\multirow{2}{*}{Perplexity} & PTB~\cite{ptb} & PPL & \multirow{2}{*}{1}\\ 
& WikiText-2~\cite{wikitext2} & PPL & \\ \hline
Calibration & C4~\cite{t5} & - & 128\\ \hline
Fine-tuning & Alpaca~\cite{alpaca} & Accuracy & 64\\ \hline
\end{tabular}
\smallskip
\footnotesize{*BS of 24 for LLaMa-2-13B}
\vspace{-12pt}
\end{table}

\subsubsection{Dataset Metrics and Evaluation} 11 natural language datasets across four tasks as shown in Table~\ref{tab:datasets} are considered. \bailey{These datasets cover a wide range of dynamic inputs and are used to evaluate model performance across diverse domains and task settings.} They are summarized as follows: 

(1) \textit{Common-sense reasoning task accuracy -} seven datasets used to assess language comprehension for a variety of reasoning tasks commonly used for evaluating LLM accuracy~\cite{openelm}. In the results presented in this article, model accuracy is reported as the equal-weighted mean across all seven datasets. Note that while some datasets present accuracy as a percentage of correct answers, others use normalized accuracy, where accuracy is adjusted according to the difficulty of the answer~\cite{openelm}. Accuracy is obtained using LM Evaluation Harness v0.4.3. A higher value indicates better performance for accuracy metrics.

(2) \textit{Perplexity -} two datasets are used to evaluate perplexity, which measures the average entropy of next-token predictions. A lower perplexity value denotes better performance, reflecting low entropy. 

(3) \textit{Calibration -} \bailey{As considered in the existing pruning literature,} 128 samples from the C4~\cite{t5} dataset are used to calculate the activations in Equation~\ref{eq:3} and Equation~\ref{eq:5}, and to calculate the inverse Hessian in SparseGPT. 

(4) \textit{Fine-tuning -} a synthetic dataset Alpaca~\cite{alpaca} is used to fine-tune a low-rank adapter (LoRA) of pruned models to recover reasoning accuracy post-pruning.

\subsubsection{Other Metrics} Accuracy is assessed based on zero-shot performance. Time and inference latency is the average from five trials, and one standard deviation is also presented. Memory use of the GPU is the allocated amount in \texttt{nvidia-smi} and for CPU is obtained from \texttt{/proc/meminfo}.

\subsection{Results}
\label{subsec:results}

The experiments evaluate the following: 

\textbf{E1} - Non-uniform pruning underpinning \system{} outperforms uniform pruning in terms of accuracy and perplexity.

\textbf{E2} - Projection pruning enables fine-grained control, leading to more optimally pruned LLM components. 

\textbf{E3} - Composite projection pruning leverages the higher accuracy of unstructured pruning with memory efficiency and faster inference of structured pruning.

\textbf{E4} - Fine-tuning LLMs after projection pruning regains accuracy faster and more effectively than uniform pruning. 

\textbf{E5} - \system{} achieves better overall performance with lower overheads than other methods. 

\textbf{Additional experiments}, such as evaluating \system{} with model quantization and reduced calibration sample size, and a full breakdown of the accuracy on individual datasets and models is provided in the Appendix.

The results presented in this article are based on analyzing over 5 TB of data produced by the experiments. 

\subsubsection{Projection Pruning Performance (E1)}
\label{e1}
Global, layer and projection pruning is evaluated on all models in Table~\ref{tab:models} on the perplexity and accuracy datasets in Table~\ref{tab:datasets}. Each model is evaluated for varying sparsities up to 80\% as the model collapses beyond this range~\cite{sparsegpt, synflow}. Perplexity is reported per dataset, whereas accuracy is the mean across the seven datasets shown in Table~\ref{tab:datasets}.

\begin{tcolorbox}[
    width=0.48\textwidth,
    colframe=black!50!black,
    colback=gray!5,boxsep=3pt,boxrule=0.5pt,left=3pt,right=3pt,top=1pt,bottom=1pt]
\textbf{Observation 1:} Projection pruning achieves up to 84.2\% lower perplexity and up to 31.4\% higher accuracy than global and layer pruning for the different models, datasets, and sparsities considered.
\end{tcolorbox}

\begin{figure*}[t]
  \centering
  \includegraphics[width=1.0\textwidth]{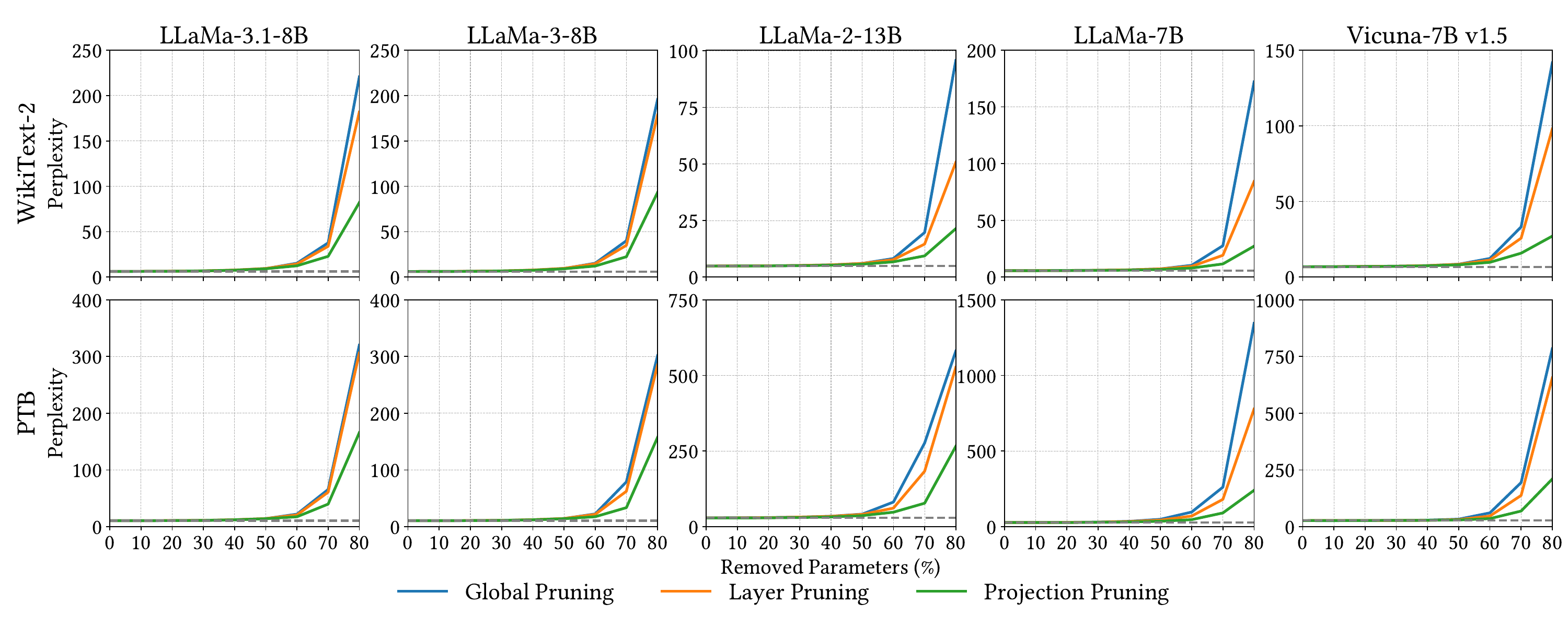}  
    \caption{Perplexity (lower is better) on WikiText-2 and PTB for LLMs as parameters are removed using different pruning methods. Dashed line is perplexity of the foundation model.}
    \label{fig:7}
\end{figure*}

\textbf{Perplexity:} Figure~\ref{fig:7} shows the WikiText-2 and PTB dataset perplexity achieved on the five LLMs for global, layer and projection pruning up to 80\% sparsity (removed parameters). As sparsity increases, perplexity naturally increases as there are fewer parameters in the LLM. Projection pruning has the lowest perplexity for all models, sparsities, and datasets. At 80\% sparsity, projection pruning ranges from 18.9\% to 84.2\% lower perplexity on WikiText-2 than global pruning and 16.8\% to 82.1\% on PTB. Layer pruning is positioned in-between projection and global pruning on older LLaMa models, such as LLaMa-7B and LLaMa-2-13B, but with newer LLaMa-3-8B and LLaMa-3.1-8B perform much closer to global pruning such as only 5.4\% lower perplexity than global on LLaMa-3-8B at 80\% sparsity on PTB.

\textbf{Accuracy:} Table~\ref{tab:acc} shows the mean zero-shot accuracy of LLaMa-3.1-8B and LLaMa-2-13B for each pruning method at 20\%, 40\%, 60\%, and 80\% sparsity. 0\% sparsity is the accuracy of the original LLM with no pruning. Projection pruning maintains the highest accuracy for both models and all sparsity values. At 40\% sparsity, projection pruning improves accuracy by less than 1\% for LLaMa-2-13B against other methods. However, at 80\% sparsity, this difference increases to 31.4\% higher accuracy. Similarly, LLaMa-3.1-8B achieves 13.2\% higher accuracy at the same sparsity level.

\begin{table}[]
\caption{Mean zero-shot accuracy of two LLMs as parameters are removed using global, layer and projection pruning.}
\label{tab:acc}
\footnotesize
\centering
\begin{tabular}{clccccc}
\hline
\multicolumn{1}{l}{}          &                          & \multicolumn{5}{c}{Removed Parameters} \\
Model                         & \multicolumn{1}{c}{Method} & 0\% & 20\% & 40\% & 60\% & 80\% \\ \hline
\multirow{3}{*}{\footnotesize{LLaMa-3.1-8B}} & Global                     &\textbf{69.35}       &68.88        &65.98       &55.53       &37.29       \\
                              &Layer&\textbf{69.35}       & 68.93        & 66.24       & 57.68       & 38.79       \\
                              & Projection                 & \textbf{69.35}      & \textbf{69.02}       &\textbf{66.41}       &\textbf{60.86}       &\textbf{42.89}       \\ \cline{1-7} 
\multirow{3}{*}{\footnotesize{LLaMa-2-13B}}  &Global                     &\textbf{67.07}       &66.12        &64.71       &59.75       &36.90       \\
                              & Layer                      &\textbf{67.07}       &66.37        &65.13       &60.95       &39.98       \\
                              & Projection                 &\textbf{67.07}       &\textbf{66.43}        &\textbf{65.31}       &\textbf{62.48}       &\textbf{48.48}       \\ \hline
\end{tabular}
\end{table}

\subsubsection{Projection Pruning Control (E2)}
\label{e2}
\begin{figure*}[t]
  \centering
  \includegraphics[width=0.98\textwidth]{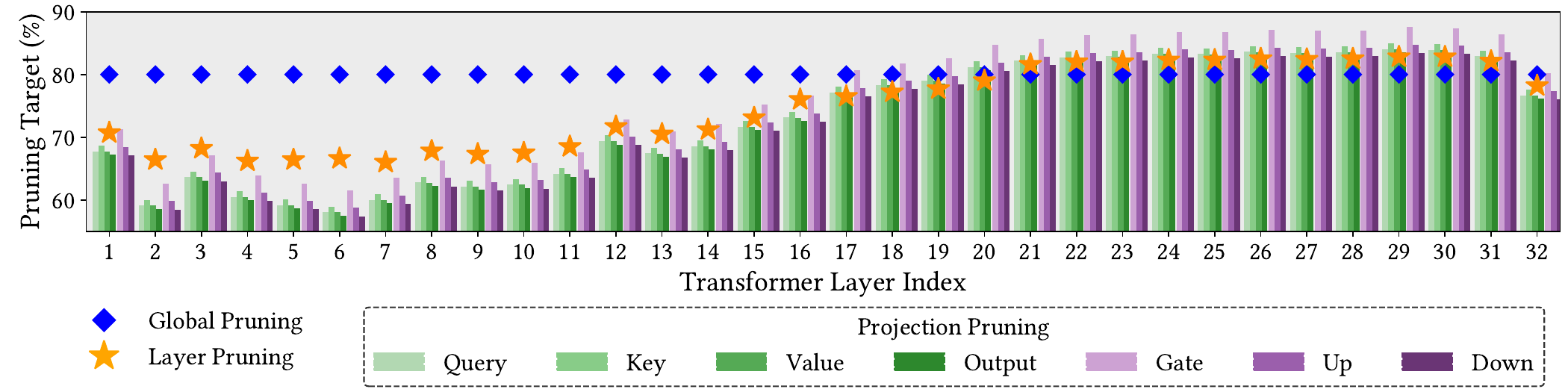}  
    \caption{Pruning targets across all layers and projections for LLaMa-3.1-8B pruned by 80\%. For projection pruning, green bars represent the attention block projections, whereas purple bars represent the feed-forward block projections.}
    \label{fig:8}
\end{figure*}

\bailey{This experiment examines the empirical differences between global, layer, and projection pruning, highlighting how the choice of granularity significantly affects pruning outcomes.} Figure~\ref{fig:8} shows the pruning target of each layer of LLaMa-3.1-8B pruned by 80\% for each method. On WikiText-2, global, layer and projection pruning achieve a perplexity of 221, 182 and 82, respectively. On PTB, 320, 306 and 166, respectively.

\textbf{Global Pruning:} Each layer is uniformly pruned as shown by the horizontal blue line in Figure~\ref{fig:8}. No distinction is made between the importance of each layer in the LLM; consequently, this method over-prunes Layers 0-18, while other methods choose to prune less. Conversely, global pruning under-prunes Layers 19-31, while other methods choose to prune more. Critical layers are over-pruned as the importance of specific parameters and layers is not considered, adversely impacting perplexity and accuracy.

\textbf{Layer Pruning:} Each layer is pruned with a different target to allow critical and redundant layers to be identified. Layer pruning identifies that layers 0-18 should be pruned less, whereas layers 19-31 can be pruned more. As a result, for an average target of 80\%, layers are pruned in the range of 66\% to 82.8\%. Although layer pruning improves perplexity by 17.6\% and 4.4\% over global pruning for WikiText-2 and PTB, respectively, it fails to identify critical layers (1-10) that more fine-grained (projection) pruning prunes less.

\textbf{Projection Pruning:} Each projection in a layer is pruned individually, resulting in a non-uniform set of pruning targets. In Figure~\ref{fig:8}, each projection in every layer is pruned by a different target. Within the attention block (green in Figure~\ref{fig:8}), the Key projection has the most redundant parameters, whereas the Output projection is pruned the least, suggesting it has more critical parameters. Within the feed-forward block (purple in Figure~\ref{fig:8}), the Gate projection is pruned more, whereas the Down projection is pruned the least. For an average target of 80\%, projections can be pruned anywhere in the range of 57.4\% to 87.5\%, improving perplexity by 55\% and 45.8\% for WikiText-2 and PTB, respectively.

Figure~\ref{fig:8} how projection-level pruning, although each pruning method ranks parameters individually, the collective importance of parameters within a single projection can lead to varying scores across different projections. Consequently, projections within the same layer may see up to a 10\% difference in pruning to accommodate their importance. This fine-grained approach contrasts global and layer-wise pruning methods, which aggressively remove crucial parameters by pruning coarse data structures, such as entire layers, into a fixed pruning target, resulting in less accurate pruned models.

\begin{tcolorbox}[
    width=0.48\textwidth,
    colframe=black!50!black,
    colback=gray!5,boxsep=3pt,boxrule=0.5pt,left=3pt,right=3pt,top=1pt,bottom=1pt]
\textbf{Observation 2:} Projection pruning enables non-uniform pruning to identify ideal pruning targets, achieving up to 63\% lower perplexity than uniform pruning.
\end{tcolorbox}

\subsubsection{Composite Projection Pruning Performance (E3)}
\label{e3}
\begin{figure*}[t]
  \centering
  \includegraphics[width=1.0\textwidth]{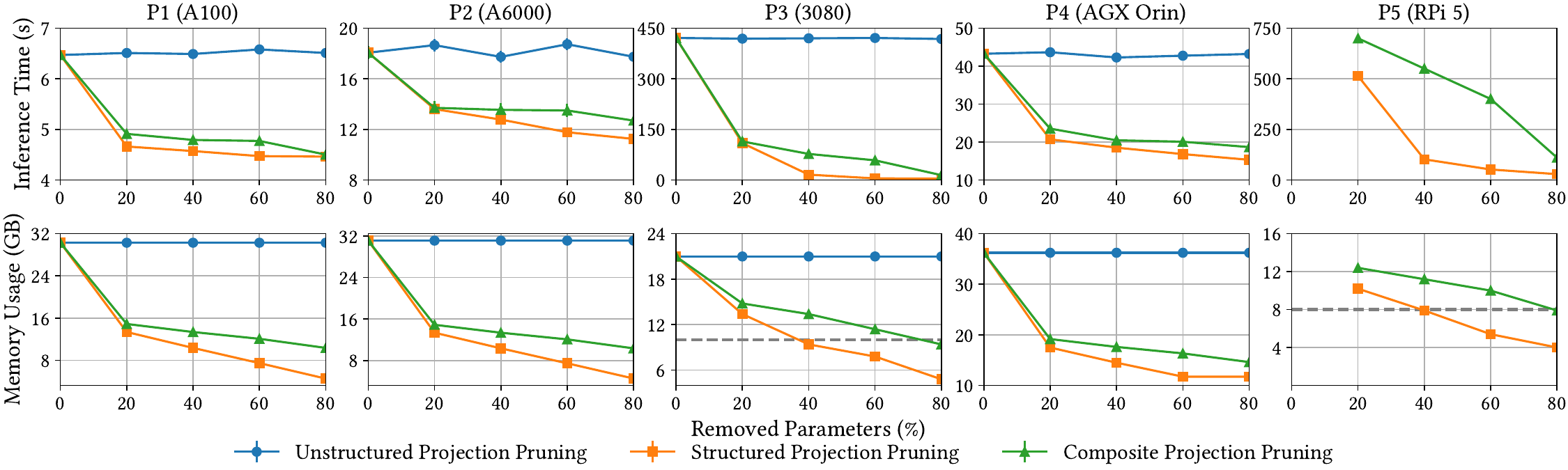}  
    \caption{Inference latency and GPU memory use of pruned LLaMa-7B for varying pruning targets (measured as \% of removed parameters) on different hardware platforms shown in Table~\ref{tab:hardware}. The dashed gray line is GPU memory capacity (for P3 and P5), and the foundation model and unstructured pruned models cannot be run on P5 due to limited resources.}
    \label{fig:sys}
\end{figure*}

\bailey{This experiment compares composite projection pruning against unstructured and structured baselines across a range of performance and accuracy metrics.} Figure~\ref{fig:sys} shows inference latency and memory usage of LLaMa-7B on the five hardware platforms shown in Table~\ref{tab:hardware} for varying pruning targets. Inference is carried out with 2048 tokens input and an output of 128 tokens per MLPerf standardized benchmark~\cite{reddi2019mlperf} with a batch size of 12. On P5, input tokens of 128 and output tokens of 16 with a batch size of 1 are used due to memory and compute constraints. Inference latency is the wall clock time for processing the entire input string and generating the output string. GPU memory used on each platform includes model weights, activations, attention, the software libraries and the \system{} framework. The size of software libraries, such as Python packages compiled for different architectures, varies across platforms, affecting the size of memory allocated. On P3 and P5 with low GPU memory, the model weights are offloaded to device storage using swap memory when exceeding GPU memory capacity (dashed gray lines in Figure~\ref{fig:sys}). Table~\ref{tab:ppl} shows the perplexity achieved on LLaMa-7B for different pruning methods.

\textbf{System Performance:} Composite and structured pruning reduce inference time and GPU memory used as more parameters are removed. Unstructured pruning does not offer any performance benefits. Although structured pruning has better runtime performance, model quality measured by perplexity (Table~\ref{tab:ppl}) is impacted. When 80\% of the parameters are removed, composite pruning has 30\% - 67\% lower inference latency than unstructured pruning while reducing memory use by 60\% - 68\%.

\textbf{Low Memory Platforms:} The GPU memory available on P3 and P5 is lower than the foundation model of LLaMa-7B. Therefore, model layers are offloaded to storage, and only a subset is loaded into memory. During inference, layers are transferred in and out of memory and storage, resulting in a data transfer bottleneck, thus increasing inference latency. For example, on P3, inference latency is 420s when using offloading but once a model is pruned enough to use less than 10GB of GPU memory, inference latency is reduced by up to 30$\times$ (14s for 80\% pruned \system{} model). The foundation model and the pruned models obtained from unstructured pruning cannot be run on P5. When the memory required is reduced from 10 GB to 7.9 GB using \system{}, the inference latency reduces from 400s to 110s. \bailey{Composite pruning of \system{} enables model deployment on memory-constrained hardware by structurally reducing the memory footprint, overcoming the limitations of unstructured pruning, which alone does not reduce inference latency or enable execution under tight memory budgets. This approach allows Mosaic to adapt the sparsification strategy to hardware constraints by applying structured pruning where unstructured sparsity alone cannot be exploited due to a lack of hardware accelerators or limited memory capacity.}

\textbf{Model Quality:} Compared to structured pruning, models derived from composite pruning have a lower perplexity as more parameters are removed. At pruning targets higher than 40\%, the models produced by structured pruning collapse (or are rendered unusable) due to very high perplexity. Models from composite pruning have up to 36$\times$ lower perplexity than structured pruning.

\begin{tcolorbox}[
    width=0.48\textwidth,
    colframe=black!50!black,
    colback=gray!5,boxsep=3pt,boxrule=0.5pt,left=3pt,right=3pt,top=1pt,bottom=1pt]
\textbf{Observation 3:} Composite projection pruning combines the merits of unstructured and structured pruning. It makes inference 67\% faster and reduces GPU memory use by 68\% compared to unstructured pruning while achieving up to 36$\times$ lower perplexity than structured pruning. 
\end{tcolorbox}

\begin{table}[t]
\caption{Perplexity of LLaMa-7B for unstructured, composite, and structured projection pruning for five pruning targets.}
\label{tab:ppl}
\scriptsize
\centering
\begin{tabular}{clrrrrr}
\hline
\multicolumn{1}{c}{}           &             & \multicolumn{5}{c}{Removed Parameters} \\
Perplexity                       & \multicolumn{1}{c}{Method}       & 0\%   & 20\%   & 40\%  & 60\%  & 80\% \\ \hline
\multirow{3}{*}{WikiText-2}     & Unstructured & 5.68       &5.76        &6.14       &7.98       &27.24       \\
                               & Composite    & 5.68      & 15.28        & 20.76       & 80.06       & 938.00       \\
                               & Structured   &5.68       &18.61        &647.20       &4,074.49       &33,586.00       \\ \cline{1-7} 
\multirow{3}{*}{PTB}    & Unstructured &27.34       &27.61        &30.87       &46.41       &240.43       \\
                               & Composite    & 27.34       & 67.68        & 100.03       & 253.45       & 1282.09       \\
                               & Structured   &27.34       &90.02        &538.65       &2,330.66       &23,447.05       \\ \cline{2-7} \hline
\end{tabular}
\end{table}

\subsubsection{Fine-tuning Performance (E4)}
\label{e4}
\begin{figure}[t]
  \centering
  \includegraphics[width=0.49\textwidth]{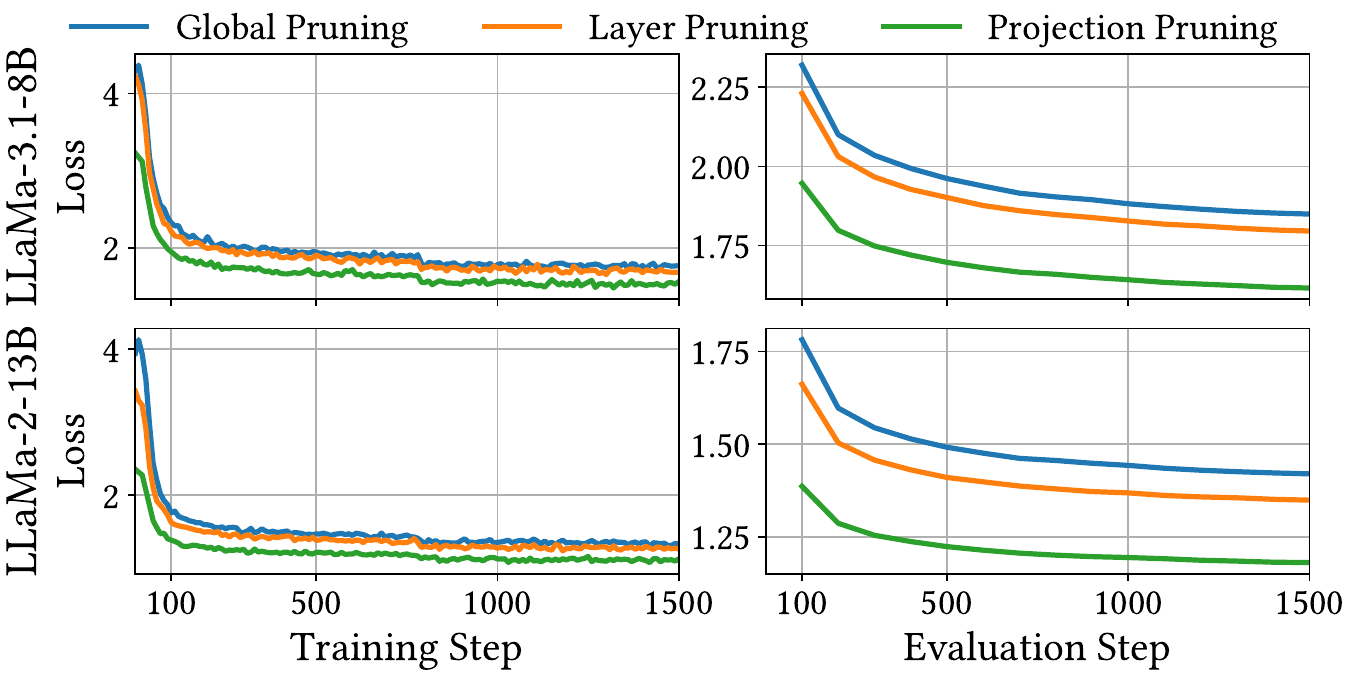}  
    \caption{Training and evaluation loss of fine-tuning an 80\% pruned LLaMa-3.1-8B and LLaMa-2-13B.
    }
    \label{fig:10}
\end{figure}

The SLM quality is regained after pruning using parameter-efficient fine-tuning (PEFT) approaches, such as low-rank adaptation~\cite{lora} (LoRA). 
Alpaca~\cite{alpaca} (Table~\ref{tab:datasets}) is used to train a LoRA adapter for LLaMa-3.1-8B for two epochs~\cite{llmpruner}. LoRA creates an 84 MB adapter which merges into the original pruned model weights at runtime. Figure~\ref{fig:10} and Table~\ref{tab:ft} summarize the findings.

\begin{tcolorbox}[
    width=0.48\textwidth,
    colframe=black!50!black,
    colback=gray!5,boxsep=3pt,boxrule=0.5pt,left=3pt,right=3pt,top=1pt,bottom=1pt]
\textbf{Observation 4:} Fine-tuning offers more performance and quality gains on LLMs compressed using projection pruning. For the same amount of fine-tuning, similar quality as global and layer pruning is achieved up to 7.5$\times$ faster, with up to 34.4\% lower perplexity and 15.4\% higher accuracy.
\end{tcolorbox}

\textbf{Training and Evaluation:} Figure~\ref{fig:10} shows the reduction in training and evaluation loss on the Alpaca dataset for each pruning method during LoRA fine-tuning. Global and layer pruning reach a final training loss of 1.76 and 1.72, whereas projection is 1.53. Projection pruning achieves a loss of 1.72 within 250 training steps in 30 minutes. This is a 6.2$\times$ speedup over other methods that need 1550 training steps and three hours to complete. Similarly, global and layer pruning reach a final evaluation loss of 1.85 and 1.80, whereas projection is 1.62. Projection reaches a loss of 1.8 at 200 evaluation steps - 7.5$\times$ faster than the other methods.

\textbf{Full Dataset Evaluation:} Table~\ref{tab:ft} demonstrates that by fine-tuning on one dataset, namely Alpaca, the perplexity and accuracy improves on other datasets listed in Table~\ref{tab:datasets}. All methods show better perplexity and accuracy following fine-tuning. Projection pruning starts with higher accuracy and gains more from fine-tuning than other methods. This indicates that projection pruning retains an optimal set of parameters that can be effectively retrained.

\begin{table}[]
\caption{Perplexity and accuracy achieved before and after fine-tuning an 80\% pruned LLaMa-3.1-8B on Alpaca.}
\label{tab:ft}
\centering
\small
\begin{tabular}{lcccc}
\hline
                          & \multicolumn{2}{c}{Before Fine-tuning} & \multicolumn{2}{c}{After Fine-tuning} \\
                        
\multicolumn{1}{c}{Method} & PPL               & Accuracy              & PPL              & Accuracy              \\ \hline
Global                     & 220.53             & 37.29             & 41.96 \tiny{($\downarrow$81.0\%)}             & 43.33 \tiny{($\uparrow$16.2\%)}             \\
Layer                      & 181.79             & 38.79             & 37.08 \tiny{($\downarrow$79.6\%)}             & 44.46 \tiny{($\uparrow$14.6\%)}             \\
Projection                 & \textbf{82.08}              & \textbf{42.89}             & \textbf{27.54 \tiny{($\downarrow$66.4\%)}}             & \textbf{50.01 \tiny{($\uparrow$16.6\%)}}             \\ \hline
\end{tabular}
\end{table}

\subsubsection{End-to-End Overheads (E5)}
\label{e5}

\begin{figure}[t]
  \centering
  \hspace*{-8pt}
  \includegraphics[width=0.51\textwidth]{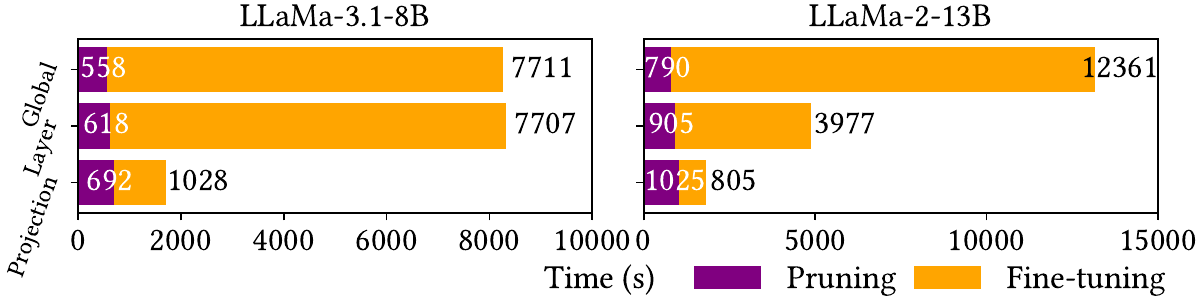}  
    \caption{End-to-end overhead (pruning time in white font and fine-tuning in black font) for LLaMa-3.1-8B and LLaMa-2-13B pruned by 80\% and fine-tuned.}
    \label{fig:bar}
\end{figure}

The combination of pruning and fine-tuning overheads is referred to as end-to-end overhead. These overheads are considered for global, layer and projection pruning in Figure~\ref{fig:bar}. 

\begin{tcolorbox}[
    width=0.48\textwidth,
    colframe=black!50!black,
    colback=gray!5,boxsep=3pt,boxrule=0.5pt,left=3pt,right=3pt,top=1pt,bottom=1pt]
\textbf{Observation 5:} The end-to-end overhead of \system{} that includes the time for projection pruning and fine-tuning to produce deployment-ready models is up to 7.19$\times$ lower than existing methods.
\end{tcolorbox}

The \textit{pruning overhead} (purple bars) of layer and projection pruning is higher than global pruning. This is because a weight metric is calculated for every parameter to obtain their pruning targets (Section~\ref{mosaic:method}), then prune the LLM based on target percentages. Global pruning of LLaMa-3.1-8B and LLaMa-2-13B requires 23.31 GB and 33.34 GB of memory, respectively, for the model parameters in FP16 precision, software libraries, 128 samples from the calibration dataset, and model activations from calibration. Layer and projection pruning requires additional memory due to the weight metrics - 24.22 GB and 35.81 GB for LLaMa-3.1-8B and LLaMa-2-13B, respectively. 

As explored in Section~\ref{e4}, LLMs are typically fine-tuned for deployment to regain accuracy lost during pruning. The \textit{fine-tuning overhead} is shown in Figure~\ref{fig:bar} (orange bars). The models pruned using layer and projection pruning are fine-tuned to match the same accuracy achieved by global pruning after it is fine-tuned for two epochs. For LLaMa-3.1-8B, projection pruning produces a deployment ready model over 4.8$\times$ faster than global and layer pruning. In the case of LLaMa-2-13B, projection pruning is 2.67$\times$ and 7.19$\times$ faster than layer and global pruning, respectively.

\section{Related Work}
\label{relatedwork}

The three main LLM compression methods to improve resource efficiency are model quantization, knowledge distillation, and pruning.

\bailey{\subsection{Model Quantization}}
State-of-the-art quantization methods, such as GPTQ~\cite{gptq} and AWQ~\cite{awq} reduce FP16 weights of the LLM into lower precision, such as INT3, INT4, or INT8, which requires less memory. Although the model size is reduced by more than half with minimal accuracy loss, the overall memory required remains high because model activations are not quantized and still use the higher FP16 precision. As memory demands grow due to larger activations from longer input lengths (more tokens), the quantized weights account for only a small fraction of the total memory usage, which reduces the overall effectiveness of this approach. Alternative methods that quantize model activations have lower accuracies and rely on a decomposition scheme to keep outlier weights at FP16~\cite{llm8}. Although all quantization methods reduce memory requirements, orthogonal approaches are required to lower inference latency. For example, GPTQ relies on custom CUDA kernels~\cite{gptq}, and AWQ utilizes QKV kernel fusion~\cite{awq}. They are, therefore, suited to GPU-based systems for performance gains, which may not be available in resource-constrained environments.

\bailey{\subsection{Knowledge Distillation}}
Smaller student models are trained to replicate the output of larger teacher models; they are used in older transformer architectures, such as BERT~\cite{bert} with DistilBERT~\cite{distilbert}. Minitron~\cite{minitron} is created using neural architecture search from Nemotron~\cite{nemotron} using pruning and knowledge distillation. Although Minitron requires 40$\times$ fewer training tokens than other similar-sized LLMs to reach the same accuracy, the computational requirements are extensive (128$\times$ Nvidia A100 80 GB GPUs~\cite{minitron}).

\bailey{\subsection{Model Pruning}}
\bailey{\textit{Unstructured Pruning:}} SparseGPT~\cite{sparsegpt} and Wanda~\cite{wanda} are \textit{unstructured pruning} methods. SparseGPT implements Optimal Brain Surgeon (OBS)~\cite{obs} and prunes models with hundreds of billions of parameters in a few hours. In contrast, Wanda employs a simpler pruning metric based solely on model weights and activations, making it two orders of magnitude faster than SparseGPT. 
Unstructured pruning does not reduce model size since parameters are set to zero. These models require specialized sparse acceleration libraries, such as NVIDIA CUTLASS, which is limited to models that are 50\% sparsity pruned using a specific \textit{semi-structured} format~\cite{sparsegpt, wanda, sparta}. 

\bailey{\textit{Structured Pruning:}} LLM-Pruner~\cite{llmpruner} and Sheared-LLaMa~\cite{sheared} are \textit{structured pruning} methods and produce smaller and faster LLMs. LLM-Pruner establishes pruning groups to efficiently prune LLM neurons in a one-shot manner, followed by fine-tuning to regain accuracy. In contrast, Sheared-LLaMa combines pruning and training in an iterative process. The pruned models are resource-efficient, but the overall model accuracy is lower than the original model. The above methods reduce layer size by removing neurons. BlockPruner~\cite{blockpruner} and ShortGPT~\cite{shortgpt} remove entire blocks and transformer layers, respectively.

\bailey{\textit{Non-uniform Pruning:}} Unlike uniform pruning that applies the same pruning ratio to all layers~\cite{sparsegpt,llmpruner}, non-uniform pruning adjusts the pruning ratio for each layer. This has better accuracy than uniform pruning for CNNs~\cite{pruning_survey, reconvene,  cai2022structured}. OWL~\cite{owl} demonstrates that this observation extends to LLMs. However, LLMs contain components within layers (projections), which in previous research have been pruned uniformly~\cite{bertpruning}. Therefore, the opportunity for non-uniform projection pruning of LLMs is explored in this article.

\bailey{\textit{Combining Unstructured and Structured Pruning for LLMs:}} Structured pruning of sparse models created by unstructured pruning has been explored for CNNs to make sparse models run on general-purpose hardware without the need for accelerators~\cite{reconvene, ECCLES202443}. 
They create small and fast sparse models. \system{} is proposed to synergistically combine both pruning methods for LLMs and apply them at the granularity of projections rather than layers or blocks for the first time, referred to as \textit{composite projection pruning}. \bailey{While existing work has explored composite pruning primarily in the context of CNNs~\cite{reconvene, ECCLES202443}, \system{} extends this approach to LLMs by targeting projection components unique to their architecture.}
\newline

\section{Conclusion}
\label{conclusion}
Deploying large language models (LLMs) on hardware-limited (compute and memory) resources remains a challenge~\cite{sparsegpt, wanda, owl}. 
Model pruning~\cite{sparsegpt, llmpruner} is one method that compresses a large foundation model to create a small language model (SLM). However, existing pruning methods negatively impact model quality~\cite{llmpruner, sheared} or rely on vendor-specific hardware and software~\cite{sparsegpt, awq}. This is because existing methods prune LLMs using coarse-grained approaches, such as uniform pruning, that inherently remove critical parameters since all layers of the LLM are pruned uniformly. Consequently, the SLMs produced are unusable when a large number of model parameters are removed~\cite{llmpruner} or require specialized software and hardware to demonstrate any performance gain when fewer parameters are removed~\cite{sparsegpt, wanda, owl}. 

This article explores a new paradigm for pruning LLMs, referred to as projection pruning. Projections are the smallest LLM components within a layer that capture intrinsic learning properties during training. Uniform pruning removes parameters from every projection equally, thereby removing parameters critical to model quality and under-pruning less important projections. While existing methods identify that specific layers should be pruned non-uniformly~\cite{owl}, this article investigates the optimization of pruning targets for every projection to maximize model quality. To this end, a novel projection-based pruning system for LLMs, \system{}, is developed to reduce the resource requirements for producing SLMs while improving model accuracy compared to other pruning methods. \system{} proposes \textit{composite projection pruning}, which combines unstructured pruning with structured pruning to produce high-quality SLMs that can be deployed on a range of hardware platforms. 
\system{} produces faster models than existing methods. \system{} models have lower perplexity and higher accuracy than coarse-grained pruning. They achieve faster inference and have a lower GPU memory use than structure pruning. 

\bailey{This work has considered dense LLM architectures. Applying projection pruning to Mixture-of-Experts (MoE) models, such as Mixtral or DeepSeek, presents a promising direction for future research. MoE architectures introduce unique challenges, such as dynamic expert routing and load balancing, that may require new pruning strategies beyond those explored here. Extending projection pruning to MoEs remains an open area for continued investigation. Additionally, many LLM pruning methods rely on calibration datasets; future work will explore developing ranking methods that remove this dependency.}

\section*{Acknowledgments}
This research is funded by Rakuten Mobile, Inc., Japan, and supported by funding from UKRI EP/Y028813/1.

\bibliographystyle{elsarticle-num}
\bibliography{references}

\appendix
\section{Hardware Platforms}
\begin{figure*}[t]
  \centering
  \includegraphics[width=0.9\textwidth]{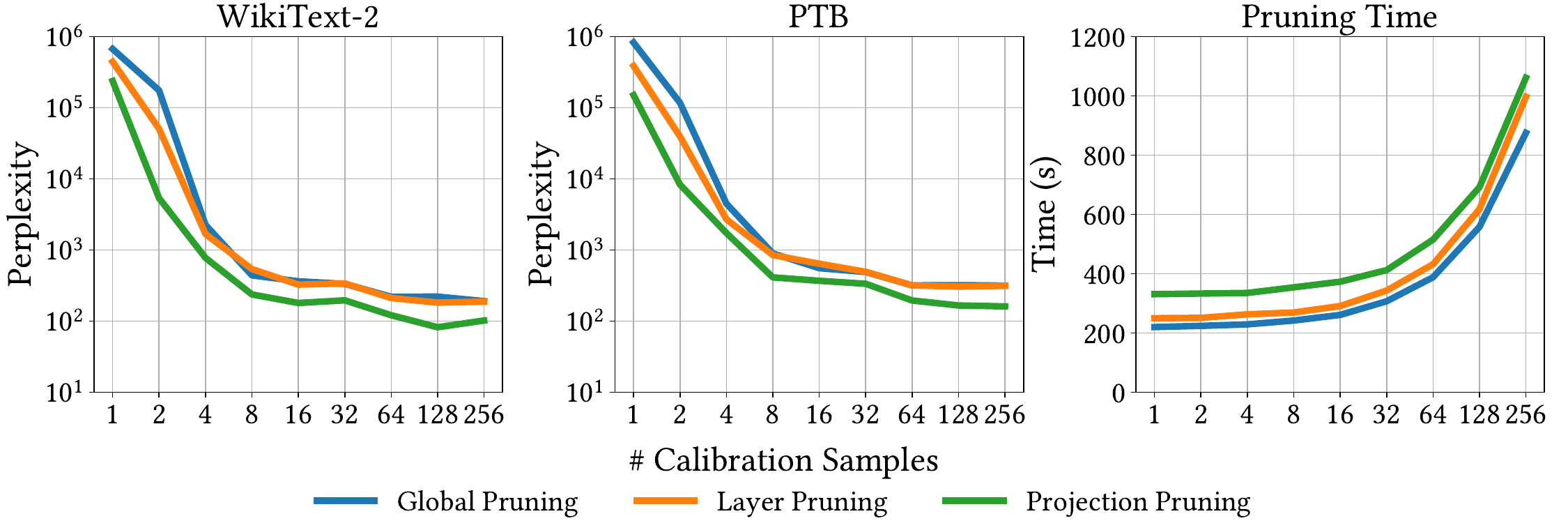}  
    \caption{Perplexity (on the WikiText-2 and PTB dataset) and pruning time for LLaMa-3.1-8B pruned by 80\% using global, layer, and projection pruning for calibration sample sizes in powers of two from 1 to 256.}
    \label{fig:samples}
\end{figure*}

Table~\ref{tab:hwd1} and Table~\ref{tab:hwd2} provide complete hardware (CPU, GPU, RAM) and OS-level breakdown of the platforms used in the experiments.
{\samepage
\section{Model Sources}
Table~\ref{tab:modelsource} provides the Hugging Face model sources for use in the Transformers Python library for each LLM used in the experiments presented in the article. We provide the exact source as the same LLM can come in different formats, such as serialized Python objects (\texttt{.bin}) or Safetensors (\texttt{.safetensors}) depending on the Transformers library version used. In addition, the same LLM may come in FP16 or FP32 bit precision, depending on the source.
\samepage
\section{Detailed Accuracy Tables}
Table~\ref{tab:verbosel31} and Table~\ref{tab:verbosel2} provide the full zero-shot accuracy breakdown for LLaMa-3.1-8B and LLaMa-2-13B results for global, layer, and projection pruning. Note that at lower pruning targets, projection pruning does not outperform all other methods; however, on average, projection pruning achieves a higher accuracy. At higher pruning targets, projection pruning demonstrates superior performance. 

\section{Calibration Sample Size}
Figure~\ref{fig:samples} shows the perplexity (on the WikiText-2 and PTB dataset) and pruning time for LLaMa-3.1-8B pruned by 80\% using global, layer, and projection pruning for calibration sample sizes $2^n$, where $n = 0, 1, 2, \cdots 8$. Most pruning methods presented in the literature choose 128 calibration samples as the default value as it optimally balances model quality against pruning time. 
Perplexity tends to improve until 128 samples, after which there is diminishing gains in perplexity with respect to pruning time. 
While projection pruning has a higher pruning time across all sample sizes, projection pruning achieves lower perplexity than global and layer pruning for all sample sizes. Notably, for half the sample size (64), projection pruning achieves a lower perplexity of 121 (WikiText-2) and 195 (PTB) than global and layer pruning at 128 samples, reaching 221 (WikiText-2, Global), 182 (WikiText-2, Global) and 320 (PTB, Global), 306 (PTB, Layer), respectively. In this case, projection pruning takes 515s, whereas global and layer pruning takes 558s and 618s, respectively.
}

\section{Comparison to Older Pruning Methods}
Table~\ref{tab:verbosel1} shows LLaMa-7B pruned by 70\% zero-shot accuracy for all seven datasets compared to older LLM pruning methods. This model and pruning target are commonly used in the literature as most methods tend to collapse beyond the 70\% pruning target, and LLaMa-7B was popular then. While more modern and better-trained LLMs are available now and were evaluated in the experiments, these tables provide a reference to compare older pruning methods.

\section{Model Quantization}
Model quantization is another compression method for LLMs that reduces the bit precision of the model parameters. This reduces the memory footprint of the model.  Table~\ref{tab:quant} compares GPTQ, a quantization method used for LLMs, for four different bit targets using the group hyperparameter of 128 for LLaMa-3.1-8B compared against \system{}. Zero-shot accuracy, as well as speedup and compression, is presented. Speedup is the improvement to LLM inference. While GPTQ can accelerate inference with custom kernels, the results presented are for hardware (P1) without the software accelerator. \system{} does not require custom kernels to accelerate inference. Compression is the file size compression of the LLM weights. While GPTQ sets weights to the target bit size, activations are still 16 bits during inference, which has the same memory usage as the unquantized/unpruned (dense) model. For \system{}, the pruning target proportionally decreases inference memory usage.

\section{Additional Calibration Datasets and Models}
\bailey{Activation-based pruning and quantization methods typically use the English subset of the C4 dataset (c4-en) and the LLaMA family of LLMs for evaluation. While C4 provides broad linguistic coverage and has been shown to generalize well across a variety of downstream tasks, it may not fully capture the needs of models trained for non-English languages. For instance, a Mistral-7B-v0.1 extended with a tokenizer designed to support Japanese characters, and fine-tuned for Japanese-language benchmarks~\cite{rakuten}. Intuitively, using a calibration dataset composed primarily of English samples may not yield a pruned model that performs optimally on Japanese reasoning tasks such as JSQuAD, JCS, JNLI, and MARC-ja~\cite{rakuten}. To evaluate this, Table~\ref{tab:japan} compares the performance of Mistral-7B-v0.1 pruned with \system{} using both the English (c4-en) and Japanese (c4-ja) subsets of C4. The model is evaluated on WikiText-2 (English) and four Japanese-language benchmarks covering a range of task types: reading comprehension (JSQuAD), multiple choice (JCS, JNLI), and text classification (MARC-ja). The results show that both calibration sets produce comparable perplexities and accuracies across all evaluation datasets. On average, c4-en slightly outperforms c4-ja in terms of mean accuracy and perplexity, particularly on JSQuAD and JNLI. However, c4-ja achieves marginally better performance on JCS and MARC-ja. These differences may be attributed to the significantly smaller size and narrower linguistic diversity of c4-ja (0.8 TB), compared to the larger and more heterogeneous c4-en dataset (10 TB), allowing for this dataset to perform well in multilingual settings.}

\begin{table*}[]
\centering 
\caption{CPU and system level details of hardware used.}
\label{tab:hwd1}
\small
\begin{tabular}{crrrrrr}
\hline
\multicolumn{1}{c}{Platform}  & \multicolumn{1}{c}{CPU} & \multicolumn{1}{c}{Cores (Threads)} & \multicolumn{1}{c}{CPU Arch.} & \multicolumn{1}{c}{Operating System} & \multicolumn{1}{c}{Kernel} & \multicolumn{1}{c}{Memory} \\ \hline
P1                                        & AMD EPYC 7713           & 64 (128) up to 3.67 GHz              & Zen 3 (x86)                                & Ubuntu 20.04.06 LTS    & 5.4.0-193-generic          & 192 GB                    \\
P2                                   & AMD EPYC 7713P          & 64 (128) up to 3.67 GHz              & Zen 3 (x86)                               &       Ubuntu 20.04.06 LTS                 &   5.15.0-117-generic                         &                  256 GB                              \\
P3                                      & Intel i9-13900KS        &  24 (32) up to 6.00 GHz                                    &    Raptor Lake (x86)                                  &   Ubuntu 20.04.06 LTS                     &   5.15.153.1-WSL2                         &  64 GB                                                \\
P4                                           & Cortex-A78AE            &         12 (12) up to 2.20 GHz                            &       ARMv8.2-A (ARM)                               &  Ubuntu 20.04.06 LTS                  & 5.10.104-tegra                          &  64 GB                                              \\
P5                                             & Cortex-A76              &      4 (4) up to 2.40 GHz                               &     ARMv8.2-A (ARM)                                 &    Debian 12 (bookworm)                   &   6.6.51+rpt-rpi-2712                         &  8 GB                                               \\ \hline
\end{tabular}
\end{table*}

\begin{table*}[]
\centering 
\caption{GPU hardware details.}
\label{tab:hwd2}
\small
\begin{tabular}{crrrrrrr}
\hline
\multicolumn{1}{c}{Platform} & \multicolumn{1}{c}{\# GPUs} & \multicolumn{1}{c}{GPU} & \multicolumn{1}{c}{GPU Memory} & \multicolumn{1}{c}{Bandwidth} & \multicolumn{1}{c}{Cores} & \multicolumn{1}{c}{GPU Arch.} & \multicolumn{1}{c}{TDP}  \\ \hline
P1                           & 2 &  Nvidia A100          &     80 GB          &   1935 GB/s                              & 6,912    &   Ampere        &   400W                  \\
P2                           &   2       &   Nvidia RTX A6000        &   48 GB            &     768 GB/s                           &     10,752                   &    Ampere                        &   300W                                             \\
P3                           &        1      &    Nvidia RTX 3080     &             10 GB                        &      760 GB/s                                &     8,704                   &       Ampere                     &     320W                                             \\
P4                           &       1          &  Jetson AGX Orin SoC        & 64 GB                         &  205 GB/s                                 &    2,048                  &  Ampere                     &            60W                                        \\
P5                           &       1          &    Broadcom BCM2712         &         4 GB                            &         15 GB/s                             &      12                  &   VideoCore VII                         &   27W                                              \\ \hline
\end{tabular}
\end{table*}

\begin{table*}[]
\centering 
\caption{Hugging Face source for each LLM used in the experiments.}
\label{tab:modelsource}
\begin{tabular}{lr}
\hline
Model        & Source    \\ \hline
LLaMa-3.1-8B                  &\texttt{meta-llama/Llama-3.1-8B}                           \\
LLaMa-3-8B                  &\texttt{meta-llama/Meta-Llama-3-8B}                          \\
LLaMa-2-13B                  &                           \texttt{TheBloke/Llama-2-13B-fp16}\\
LLaMa-7B                  & \texttt{huggyllama/llama-7b}                           \\
Vicuna-7B v1.5                  &\texttt{lmsys/vicuna-7b-v1.5}                           \\\hline                        
\end{tabular}
\end{table*}

\begin{table*}[]
\centering 
\caption{LLaMa-3.1-8B zero-shot accuracy on the seven datasets using global, layer, and projection pruning for different pruning targets. Bold values denote the highest accuracy for each pruning target.}
\label{tab:verbosel31}
\begin{tabular}{clcccccccc}
\hline
Pruning Target        & Method                      & ARC-c & ARC-e & BoolQ & HellaSwag & OBQA & RTE   & WinoGrande & Mean \\ \hline
o\%                   & -                           & 53.50  & 81.19 & 82.02 & 78.85     & 44.60 & 71.84 & 73.48      & 69.35 \\ \hline
\multirow{3}{*}{20\%} & Global  & 52.65 & \textbf{80.85} & 81.83 & 79.00        & 44.80 & 69.68 & 73.32      & 68.88 \\
                      & Layer         & 53.07 & 80.22 & 81.87 & \textbf{79.18}     & 44.40 & \textbf{70.04} & \textbf{73.72}      & 68.93 \\
                      & Projection & \textbf{53.16} & 80.39 & \textbf{82.14} & 79.00        & \textbf{45.20} & 69.68 & 73.56      & \textbf{69.02} \\ \hline
\multirow{3}{*}{40\%} & Global  & \textbf{50.43} & \textbf{77.02} & 78.23 & \textbf{77.38}     & \textbf{43.80} & 62.09 & 72.93      & 65.98 \\
                      & Layer        &   50.09    &     75.67  &  79.88     &   76.65        & 42.80     &   65.34    &    \textbf{73.24}        &  66.24     \\
                      & Projection & 50.34  & 75.34 & \textbf{79.94} & 76.77 & 42.60 & \textbf{67.87} &    71.98        & \textbf{66.41}       \\ \hline
\multirow{3}{*}{60\%} & Global  &  34.73     &  57.32    &   75.29    &    62.12       &  37.40    & 54.15      &   67.72        &  55.53     \\
                      & Layer         &    37.54   &  61.99     &  77.25     &   64.35        & 37.60     &   56.68    &    68.35        & 57.68      \\
                      & Projection &  \textbf{42.23}     &   \textbf{67.80}    &   \textbf{78.92}    &   \textbf{68.72}        &  \textbf{38.80}    &  \textbf{59.21}     &    \textbf{70.32}        &   \textbf{60.86}    \\ \hline
\multirow{3}{*}{80\%} & Global  &  20.82     &    29.42   &   52.11    &    28.43       &  27.00    &   52.35    &     50.91       &   37.29    \\
                      & Layer         &   20.14    &    30.35   &  61.47     &     30.20      &   26.80   &  \textbf{52.71}     &     49.88       &  38.79     \\
                      & Projection &   \textbf{24.49}    &  \textbf{36.66}     & \textbf{64.68}      &   \textbf{38.59}        &  \textbf{27.80}    &  \textbf{52.71}     &   \textbf{55.33}         &  \textbf{42.89}     \\ \hline
\end{tabular}
\end{table*}

\begin{table*}[]
\centering 
\caption{LLaMa-2-13B zero-shot accuracy on the seven datasets using global, layer, and projection pruning for different pruning targets. Bold values denote the highest accuracy for each pruning target.}
\label{tab:verbosel2}
\begin{tabular}{clcccccccc}
\hline
Pruning Target        & Method                      & ARC-c & ARC-e & BoolQ & HellaSwag & OBQA & RTE   & WinoGrande & Mean  \\ \hline
o\%                   & -                           &  49.15 & 77.53 & 80.55 &  79.39    & 45.20 & 65.34 &  72.30     & 67.07 \\ \hline
\multirow{3}{*}{20\%} &  Global  & 49.74  & 77.02 & 81.10 & 76.69    &  \textbf{46.20} & \textbf{59.57} &  72.53      &  66.12 \\
                      & Layer        & \textbf{49.83}   & \textbf{77.06}    & \textbf{81.22}  & \textbf{79.86}       &  45.40   &  59.21  &    71.98     &  66.37    \\
                      & Projection & \textbf{49.83}  & 76.81 & 80.70  & 79.64  & 45.80 & \textbf{59.57} & \textbf{72.69}         & \textbf{66.43}       \\ \hline
\multirow{3}{*}{40\%} & Global  & 48.04 &74.07 &81.28  &  77.59    & 45.20 &54.51 &  \textbf{72.30}     & 64.71 \\
                      & Layer        & \textbf{48.72}    &     \textbf{75.67}  & \textbf{80.61}     &   \textbf{78.24}        & 44.60     &  56.32    &    71.74        &  \textbf{65.13}     \\
                      & Projection & 47.44 & 73.70 & 81.13 & 77.65 & \textbf{46.20} &\textbf{59.21} &  71.82         & 65.31       \\ \hline
\multirow{3}{*}{60\%} &  Global  & 40.78  & 64.90 & 79.48 & 67.72    & 40.60 & 53.79 & 70.96      & 59.75 \\
                      & Layer        & 40.53   & 64.94    &81.16   &   68.68     &  41.00   & \textbf{58.84}   &  \textbf{71.51}      &  60.95    \\
                      & Projection & \textbf{43.60}  & \textbf{69.74} & \textbf{81.93}  & \textbf{72.80}  &  \textbf{42.00} & 56.32 & 70.96         & \textbf{62.48}       \\ \hline
\multirow{3}{*}{80\%} & Global  & 22.95 & 28.91 & 51.25 &   29.64   & 24.60 &52.71 &   48.22    &36.90  \\
                      & Layer        &  21.76    & 32.53  & 62.17     &   31.39        & 27.20     & 52.71    & 52.09        & 39.98     \\
                      & Projection & \textbf{27.90} &\textbf{46.09}  & \textbf{68.38} & \textbf{47.47} & \textbf{32.60} &\textbf{53.79} &    \textbf{63.14}       &   \textbf{48.48}     \\ \hline                      
\end{tabular}
\end{table*}

\begin{table*}[]
\centering 
\caption{LLaMa-7B pruned by 70\% zero-shot accuracy on seven datasets for different pruning methods.}
\label{tab:verbosel1}
\begin{tabular}{clcccccccc}
\hline
Pruning Target        & Method                      & ARC-c & ARC-e & BoolQ & HellaSwag & OBQA & RTE   & WinoGrande & Mean  \\ \hline
o\%                   & -                           & 41.38  & 67.67 & 75.14  &  74.80    & 41.40 & 66.43 &  70.01     & 62.40 \\ \hline
\multirow{5}{*}{70\%} & Magnitude  & 22.35  &  26.98 & 38.29 & 24.68    &  25.80 & 52.71 & 51.46      &  34.61 \\
                      & Wanda        &19.80    & 34.22    & 55.11  &   31.83     & 26.00   & 57.40   & 51.38    & 39.39    \\
                      & SparseGPT & 24.57  & 43.06 & 64.53 & 42.11  &  27.80 & 53.79 & 58.64 & 44.93 \\
                       & OWL        &27.65    & 45.41    & 67.13  & 48.56       & 32.00   & 53.43   &  62.03    & 48.03    \\
                      & \system{} & \textbf{30.63}  & \textbf{49.45}& \textbf{69.72} & \textbf{54.72}  & \textbf{35.80} & \textbf{58.84} & \textbf{64.33}         &  \textbf{51.93}\\ \hline                
\end{tabular}
\end{table*}

\begin{table*}[]
\centering 
\caption{Model quantization and pruning zero-shot accuracy, speedup, and compression (comp.) of LLaMa-3.1-8B for different bit and pruning targets.}
\footnotesize
\label{tab:quant}
\begin{tabular}{cccccccccccc}
\hline
Category        & Target                      & ARC-c & ARC-e & BoolQ & HellaSwag & OBQA & RTE   & WinoGrande & Mean  & Speedup  & Comp. \\ \hline
Dense                  & 16 bit / 100\%                          & 53.50  & 81.19 & 82.02 & 78.85     & 44.60 & 71.84 & 73.48      & 69.35  & 1.00$\times$      & 1.00$\times$ \\ \hline
\multirow{4}{*}{Quantization (GPTQ)} & 8 bit  & 53.24  & 77.61 & 81.28 &   79.12    &  44.80 & 68.95 &  72.85      &  68.26 & 0.48$\times$ &  1.74$\times$ \\
                      & 4 bit        & 38.40   &  64.81   & 79.39  &  76.24      &  41.80  &  66.43  &  71.59   & 62.67 & 0.47$\times$&2.80$\times$   \\
                      & 3 bit & 23.89  & 36.32& 55.05 & 39.29  & 31.20 & 51.99 & 55.56        & 41.90 & 0.44$\times$ &  3.31$\times$\\
                       & 2 bit        &26.45    &24.83     &49.69   &26.14      &27.40   &48.38    & 48.22   &35.87   &0.33$\times$ &4.04$\times$  \\ \hline      
\multirow{4}{*}{Pruning (\system{})} & 20\%   & 53.16 & 80.39 & 82.14 & 79.00        & 45.20 & 69.68 & 73.56      & 69.02  &  1.32$\times$&  1.24$\times$ \\
                      & 40\%       & 50.34  & 75.34 & 79.94 & 76.77 & 42.60 & 67.87 &    71.98        & 66.41  &1.35$\times$ &1.59$\times$   \\
                      &  60\% & 42.23     & 67.80    & 78.92    & 68.72        & 38.80   & 59.21     & 70.32        &   60.86 &  1.36$\times$ & 2.33$\times$\\
                       & 80\%       & 24.49    &  36.66     & 64.68      &   38.59        &  27.80    &  52.71     &   55.33         &  42.89   & 1.44$\times$& 4.20$\times$ \\ \hline                         
\end{tabular}
\end{table*}

\begin{table*}[]
\centering 
\caption{\bailey{Mistral-7B-v0.1 pruned by 50\% zero-shot accuracy and perplexity on four Japanese datasets for two different C4 calibration datasets.}}
\label{tab:japan}
\begin{tabular}{c|c|cccc|c}
\hline
\bailey{Calibration Dataset}     & \bailey{WikiText-2} & \bailey{JSQuAD} & \bailey{JCS} & \bailey{JNLI} & \bailey{MARC-ja} & \bailey{Mean}  \\ \hline
\bailey{- (Base Model)}  & \bailey{5.47} & \bailey{78.97} & \bailey{84.18} & \bailey{56.82}     & \bailey{96.39} & \bailey{79.09} \\ \hline
\bailey{c4-en}      & \bailey{\textbf{7.21}} & \bailey{\textbf{72.11}} & \bailey{68.01} & \bailey{\textbf{29.87}}     & \bailey{89.12} & \bailey{\textbf{64.78}}  \\
\bailey{c4-ja}  & \bailey{7.26} & \bailey{70.08} & \bailey{\textbf{69.52}} & \bailey{27.65}     & \bailey{\textbf{91.50}} & \bailey{64.69}  \\ \hline
\end{tabular}

\end{table*}
\end{document}